\documentclass[lettersize,journal]{IEEEtran}
\usepackage{amsmath,amsfonts}
\usepackage{algorithmic}
\usepackage{algorithm}
\usepackage{array}
\usepackage[caption=false,font=scriptsize,labelfont=sf,textfont=sf]{subfig}
\usepackage{textcomp}
\usepackage{stfloats}
\usepackage{url}
\usepackage{verbatim}
\usepackage{graphicx}
\usepackage{cite}
\usepackage{comment}
\usepackage{color}

\definecolor{gray}{rgb}{0.5,0.5,0.5}
\definecolor{green}{rgb}{0, 0.6, 0}
\definecolor{orange}{rgb}{1, 0.5, 0} 	
\definecolor{mahogany}{rgb}{0.75, 0.25, 0.0}
\definecolor{purple}{rgb}{0.6, 0, 0.6}
\definecolor{blue}{rgb}{0, 0, 0.6}

\newcommand{\ignore}[1]{}

\begin{document}

\title{Screentone-Aware Manga Super-Resolution Using DeepLearning}

\author{\IEEEauthorblockN{Chih-Yuan Yao,}
\and
\IEEEauthorblockN{Husan-Ting Chou,}
\and
\IEEEauthorblockN{Yu-Sheng Lin}
\and
\IEEEauthorblockN{Kuo-wei Chen}

\thanks{.}
\thanks{.}}

\markboth{.}%
{Shell \MakeLowercase{\textit{et al.}}: A Sample Article Using IEEEtran.cls for IEEE Journals}


\maketitle

\begin{abstract}
Manga, as a widely beloved form of entertainment around the world, have shifted from paper to electronic screens with the proliferation of handheld devices. However, as the demand for image quality increases with screen development, high-quality images can hinder transmission and affect the viewing experience. Traditional vectorization methods require a significant amount of manual parameter adjustment to process screentone. Using deep learning, lines and screentone can be automatically extracted and image resolution can be enhanced.
Super-resolution can convert low-resolution images to high-resolution images while maintaining low transmission rates and providing high-quality results. However, traditional Super Resolution methods for improving manga resolution do not consider the meaning of screentone density, resulting in changes to screentone density and loss of meaning.
In this paper, we aims to address this issue by first classifying the regions and lines of different screentone in the manga using deep learning algorithm, then using corresponding super-resolution models for quality enhancement based on the different classifications of each block, and finally combining them to obtain images that maintain the meaning of screentone and lines in the manga while improving image resolution.
\end{abstract}

\section{Introduction}
In order to display high-resolution images of manga on different display devices while maintaining low transmission bandwidth, the most common traditional method is to convert them into vector graphics then calculate the corresponding pixel colors based on the kernel function of vectorization, when display on different devices. However, according to the research by Yao et al.\cite{7399427}, a large amount of manual operation is required to adjust the parameters for both the segmentation of screentone and the vectorization results. In order to achieve less data transmission and high-resolution manga images more quickly and automatically, deep learning is used to replace a large number of parameter adjustments by retraining the deep learning network with a large amount of pre-annotated screentone data to automatically segment the areas, and the super-resolution algorithm is used to convert low-resolution images into high-resolution ones.

\begin{figure}[htbp]
    \centering
    \subfloat[Original image]{
        \includegraphics[scale=0.3]{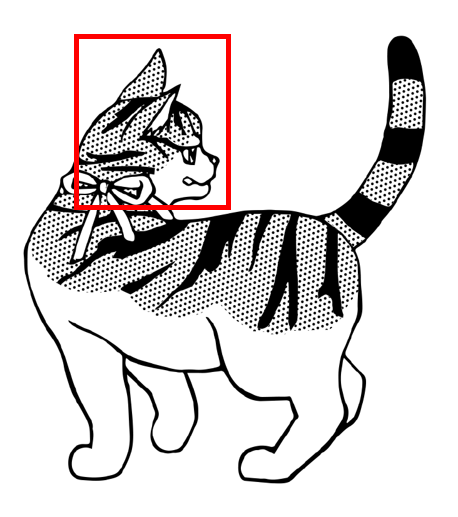}
    }
    \subfloat[Upscale by super-resolution]{
        \includegraphics[scale=0.3]{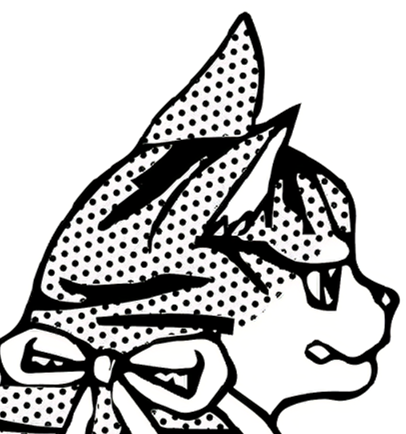}
    }
    \subfloat[Expect result after upscale]{
        \includegraphics[scale=0.3]{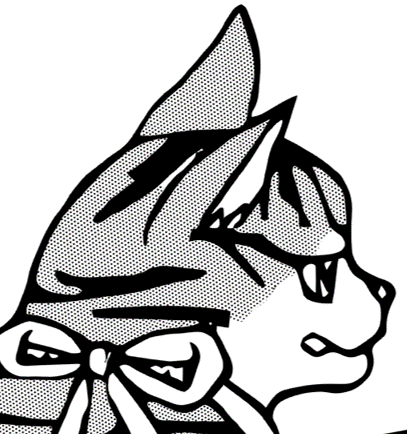}
    }
    \caption{Traditional super-resolution algorithm cannot maintain the density of manga screen tones when scaling images, resulting in the loss of the original meaning represented by the screen tones.}
    \label{CH1_fig1}
\end{figure}

Super-resolution can convert images from low-resolution to high-resolution, but traditional methods only focus on maintaining the content of the images, without considering the preservation of the image meaning. The screentone in manga originate from the limitations of printing technology and are used to create gray tones through visual illusions. The color depth of the area can be controlled based on the density, and manga artists use this technique to give images different effects. However, such effects can be destroyed when using traditional super-resolution methods, as shown in figure~\ref{CH1_fig1}. Because the density of dot blocks was not considered when converting to high resolution, areas that should have been uniform gray became many black screentone in figure~\ref{CH1_fig1}(a). The correct approach, as shown in figure~\ref{CH1_fig1}(b), is to maintain the original density and reapply the screentone.

The main purpose of this study is to apply super-resolution algorithms to increase the resolution of manga images according to demand while maintaining the meaning represented by the screentone in the manga as much as possible. Firstly, we adjust the Super Resolution method based on deep learning, as a single training model cannot be applied to different types of screentone, we use a semantic segmentation network to classify the screentone contained in the image, and then use corresponding Super Resolution networks to enlarge them separately. At the same time, we propose a new loss function to deal with the artifacts generated during the generation of some structured screentone. Finally, we will verify whether the density of the restored screentone is the same as that of the input low-resolution image. Our main contributions are as follows:
\begin{itemize}
\item We propose a method to automatically maintain the density of screentone when enlarging comics, instead of simply enlarging them.
\item Compared to vectorization methods, our method reduce manual operation and can handle more complex screentone.
\item We propose a new loss function to deal with the artifacts generated during the generation of some structured screentone, resulting in better results.
\end{itemize}


\section{Related works}
In order to magnify a low-resolution image, we focus on deep neural network approaches to solve the problem.  However, it gives rise to several problems such as screentone misidentified, distortion. To solve these problems,  we decided to apply image semantic segmentation and repeat pattern detection to classify screentone and discriminate screentone structure. These problems are examined in detail in latter chapters.

\subsection{Super Resolution using Deep Learning}
Super resolution refers to the process of using software or hardware to convert a low-resolution image (LR) into a high-resolution image (HR). How to preserve as much detail as possible when upsampling is a key challenge in super resolution processing. With deep learning techniques, super resolution preserves more detail information in images and has better results in terms of image quality assessment such as peak signal-to-noise ratio (PSNR). Dong et al. \cite{DBLP:journals/corr/DongLHT15} first applied deep learning to the field of super-resolution, and Kim et al. \cite{DBLP:journals/corr/KimLL15b} deep the network structure and introduced the residual network ResNet to solve the problem of training deeper networks. Both deep learning networks above used pixel-based loss functions and achieved better results in PSNR and SSIM. However, the restored results still appeared too smooth in some areas with high-frequency details.

Ledig et al. \cite{DBLP:journals/corr/LedigTHCATTWS16} were the first to apply Generative Adversarial Nets (GANs)\cite{GAN} to the super-resolution problem, proposing SRGAN. The authors added a discriminative model to the SRResNet and introduced perceptual loss and adversarial loss to address the problem of using mean squared error (MSE) as a loss function, which results in a higher peak signal-to-noise ratio (PSNR) but loses high-frequency details, while also improving the realism of the restored images. Subsequent research \cite{ESRGAN, yuan2018unsupervised,  zhang2019ranksrgan, wang2021realesrgan} has made improvements to the network architecture, feature extraction block, and the design of perceptual loss in order to enhance the resolution of image restoration and make the restored details more in line with human visual perception.

In this paper, We referenced ESRGAN \cite{ESRGAN} and used its Residual-in-Residual Dense Block (RRDB) structures to help us maintain the details and density of screentones while improving the resolution of manga images. At the same time, we noticed that ESRGAN may produce artifacts or irregular spacing in screentones with stronger structural characteristics, so we propose a loss function to maintain the structural integrity of screentone.

\subsection{Manga Vectorization}
Vector image is often used to present manga images when have requests to scale for different display devices. Zhang et al. \cite{4745633} and Kopf \cite{ComicVector} segmented the image into multiple blocks using color, and vectorized each block into a gradient patch. However, high-frequency details such as screentones are lost and become smooth. Yao et al. \cite{7399427} addressed the screentone problem by separated managa image into screentones and borders. The borders were converted into Bezier curves, while the screentones were split into smaller units such as dots, stripes, and grids. Then, the screentone regions were reassembled using parameterization. However, this approach required a large number of manual parameter settings and complex patterns were difficult to define mathematically. To address this issue, we introduce deep learning techniques to automatically generate various styles of screentones with synthetic data. This approach significantly reduces the need for manual parameter settings.

\subsection{Repeat Pattern Detection}
Lettry et al. \cite{7926596} resorted to neural networks to conduct structural analysis on an image’s repeated regions. They begin by using AlexNet to obtain several feature maps that are then analyzed through Hough voting and Lattice detection voting, in turn obtaining the structure of repeated regions. In this study, we want to adopt such output structure for our analysis of repeated regions in screentone to reduce artifacts that result from super resolution when restoring certain structural screentone. 

The highlight of our study is how our proposed method enables automatic retention of screentone density when comics are magnified. Compared to the aforementioned approach, our method does not require manual intervention in the comics manipulation process; moreover, when dealing with more complex screentone, our system will simultaneously analyze the screentone structure to help achieve better results in screentone generation.

\begin{figure*}[htb]
    \centering
    \includegraphics[width=2\columnwidth]{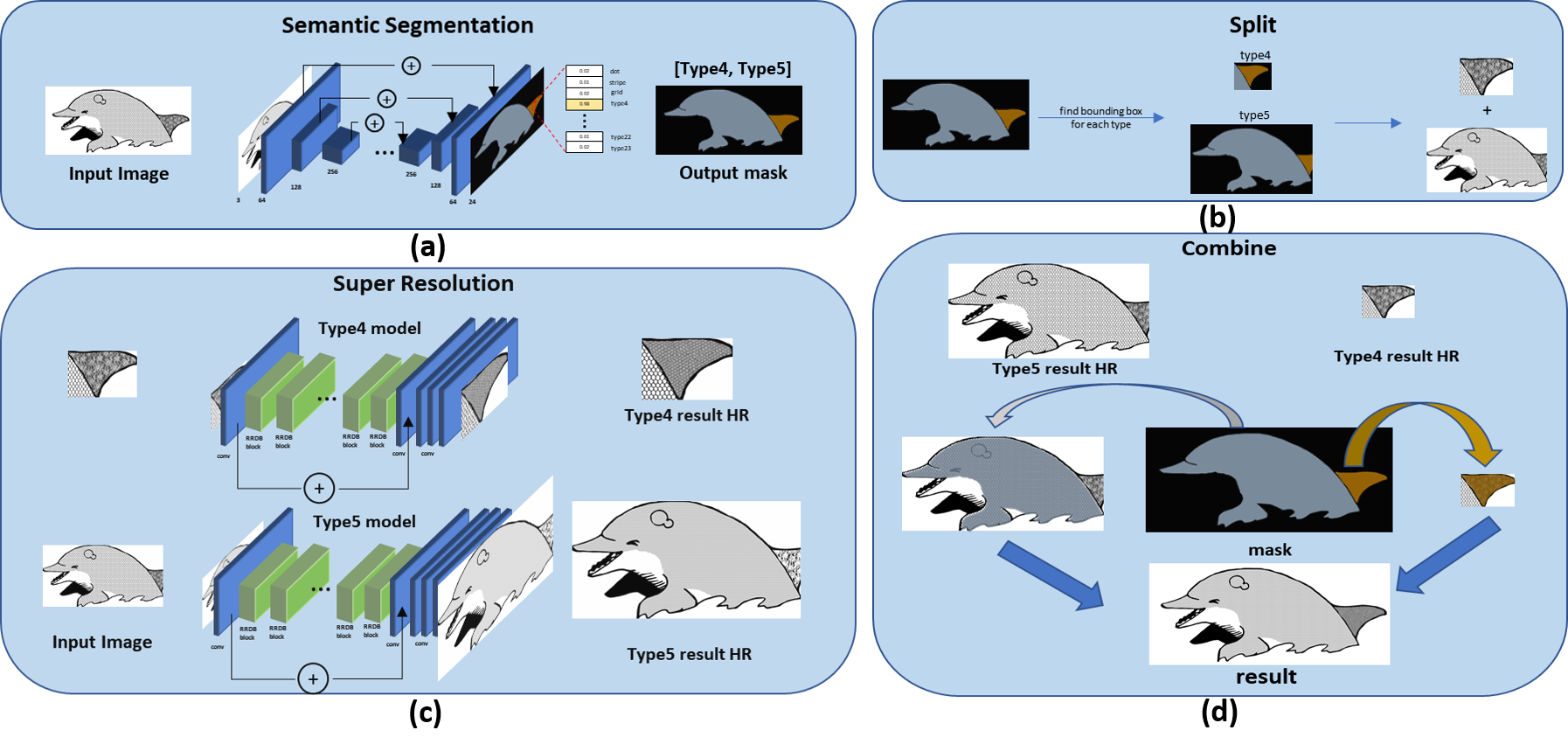}        
    \caption{System Overview. (a). First, a U-Net-based semantic segmentation model~\cite{DBLP:journals/corr/RonnebergerFB15} to segment the screentone areas of the input image and classify them into different types. (b). Each segmented area is cropped and treated as input. (c). For each cropped area, a different pre-trained super-resolution (SR) network is applied based on its corresponding screentone classification. (d). Finally, the SR results of each area are synthesized with the SR result of the original input image, resulting in an output image with enhanced resolution.}
    \label{CH3_fig1}
\end{figure*}

\section{System Overview}
The flowchart of our system is presented in Figure~\ref{CH3_fig1}. The workflow includes inputting an LR image that undergoes semantic segmentation for screentones classification, such as dots, grids, blocks, and other types of screentones. This step also generates mask images of the detected screentone in different image regions. Based on the screentone pattern detected on the LR image, the system identifies the corresponding region of that pattern. Different pre-train models are then applied to each segmented partition from the previous step to perform super resolution on the LR image. The result is one or more upscale images which maintain the density of the screentone. Finally, the system uses the mask images obtained from semantic segmentation to locate the corresponding upscale screentone region on the image which apply super resolution and render the final output.

\section{Proposed Methods}

\subsection{Image Semantic Segmentation}
\begin{figure}[htbp]
  \begin{center}
    \subfloat[original manga]{
        \includegraphics[scale=0.14]{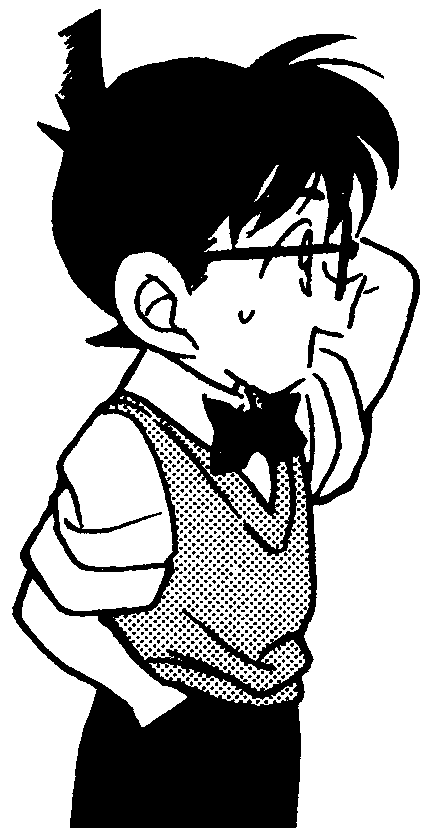}
        }
    \subfloat[line drawings]{
        \includegraphics[scale=0.14]{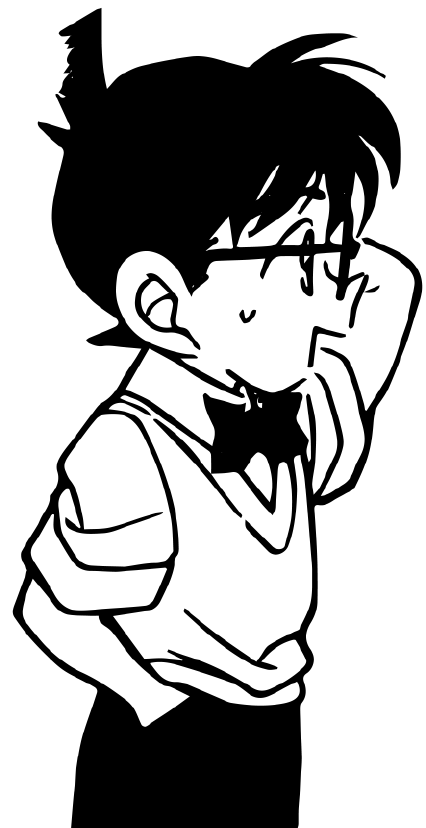}
        }
    \subfloat[mask]{
        \includegraphics[scale=0.14]{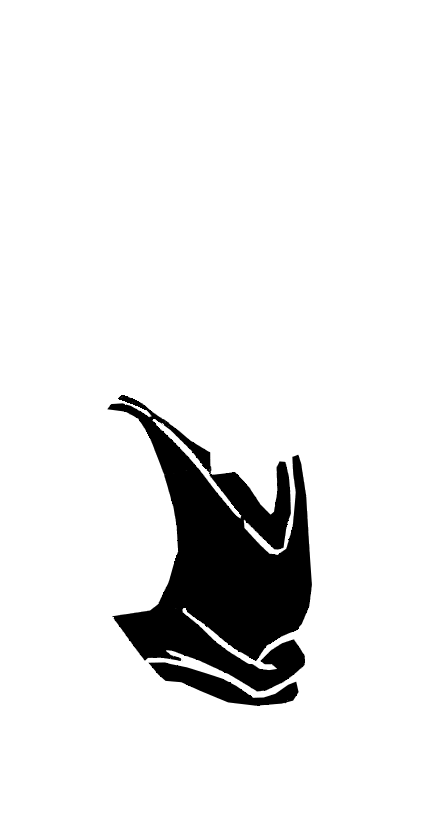}
        }
    \\
    \subfloat[original manga]{
        \includegraphics[scale=0.08]{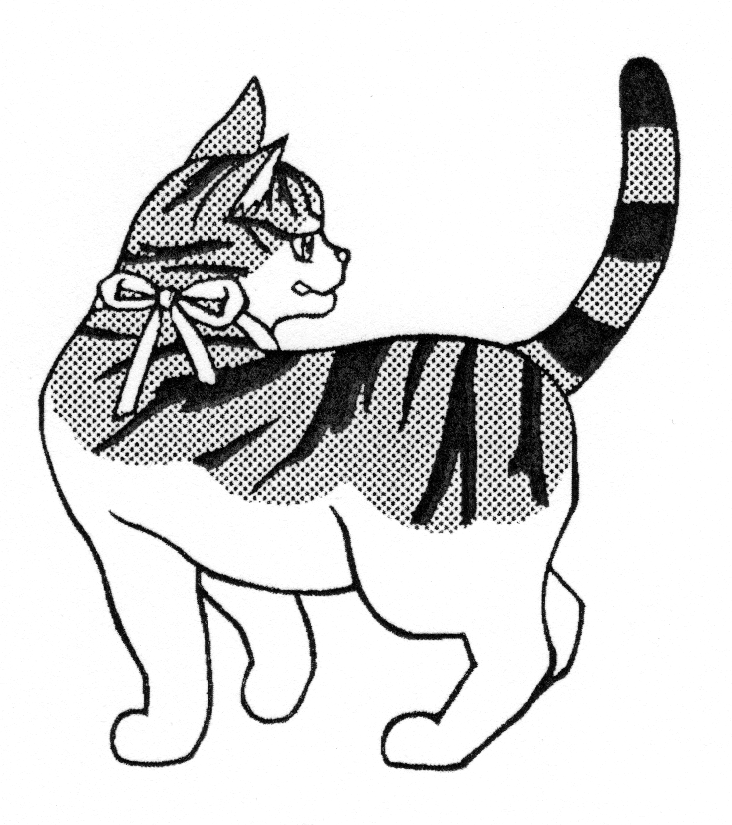}
        }
    \subfloat[line drawings]{
        \includegraphics[scale=0.08]{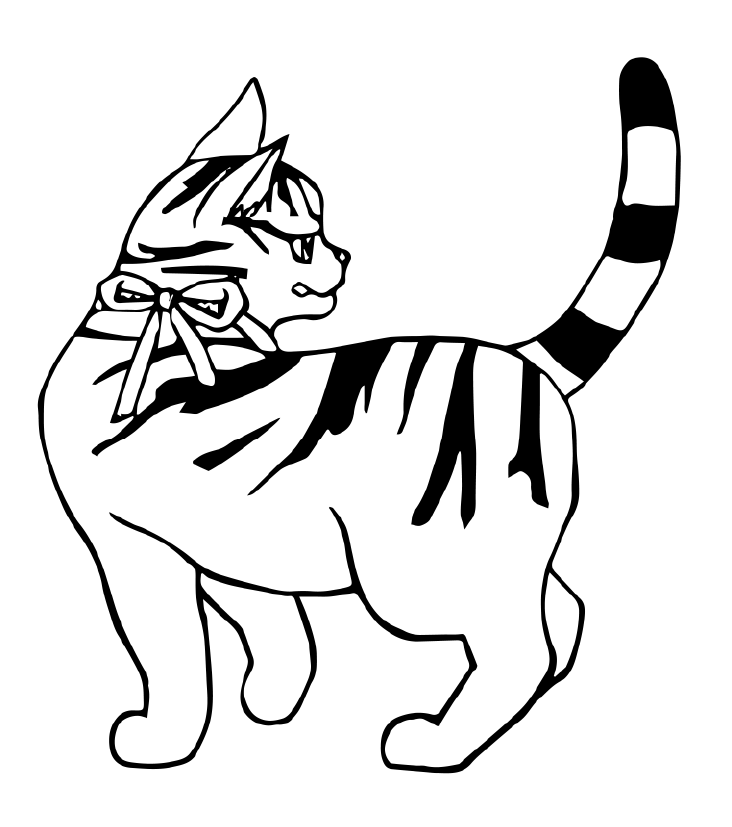}
        }
    \subfloat[mask]{
        \includegraphics[scale=0.08]{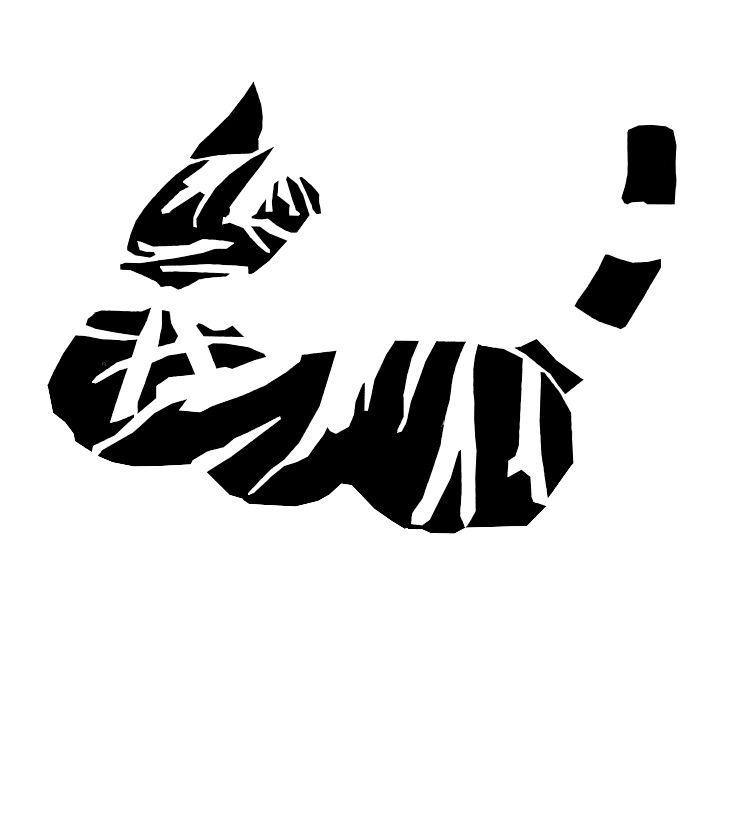}
        }
    \caption{Extract the sceentone areas.}
    \label{CH4_Sec2_fig1}
  \end{center}
\end{figure}

In this study, U-Net~\cite{DBLP:journals/corr/RonnebergerFB15} was used for semantic segmentation of mangas to detect screentone patterns and their corresponding regions. Manga line drawings filled with various screentone patterns were used as training data. Around 30 pieces of manga line drawings were collected from paper comic books and Internet resources. For paper manga, we marked the regions that were originally screentone-filled to produce a mask. For Internet-obtained manga, we marked the screentone-filled regions to produce a corresponding mask. Figure~\ref{CH4_Sec2_fig1} illustrates the process, where Figure~\ref{CH4_Sec2_fig1}(a) and (d) show the original manga, (b) and (e) show the line drawings, and (c) and (f) show the masks, respectively.

\begin{figure}[htbp]
  \begin{center}
    \subfloat[line drawings]{
        \includegraphics[scale=0.14]{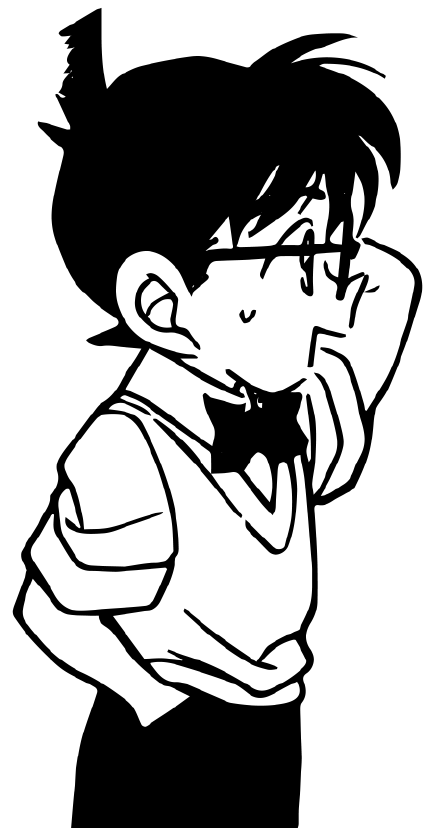}
        }
    \subfloat[screentone-filled]{
        \includegraphics[scale=0.14]{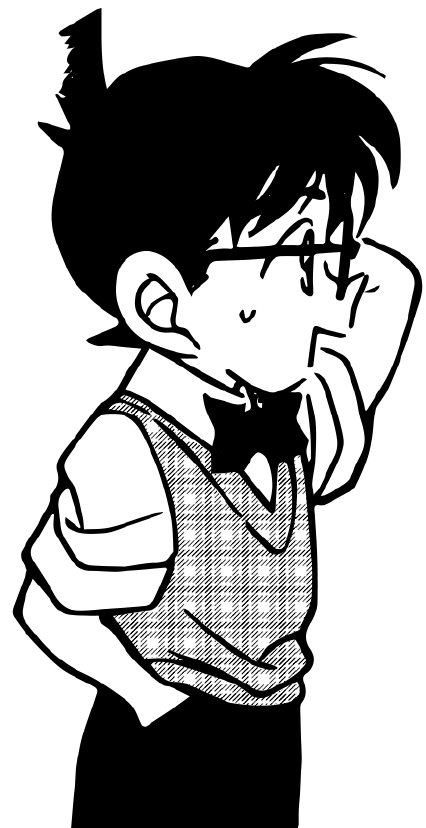}
        }
    \subfloat[mask]{
        \includegraphics[scale=0.14]{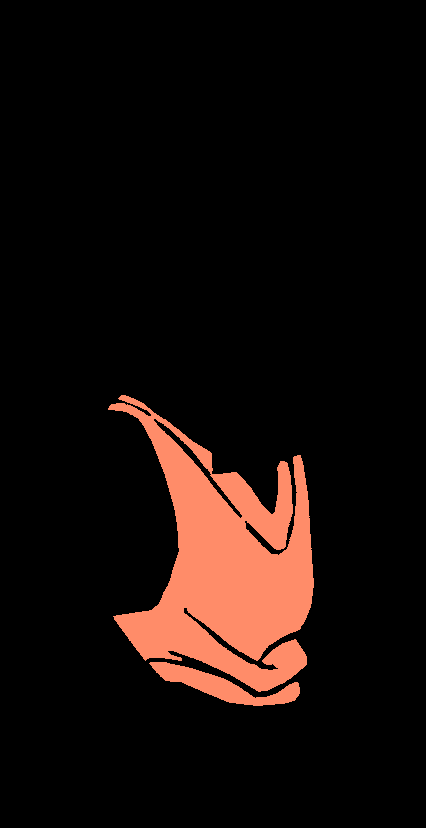}
        }
    \\
    \subfloat[line drawings]{
        \includegraphics[scale=0.065]{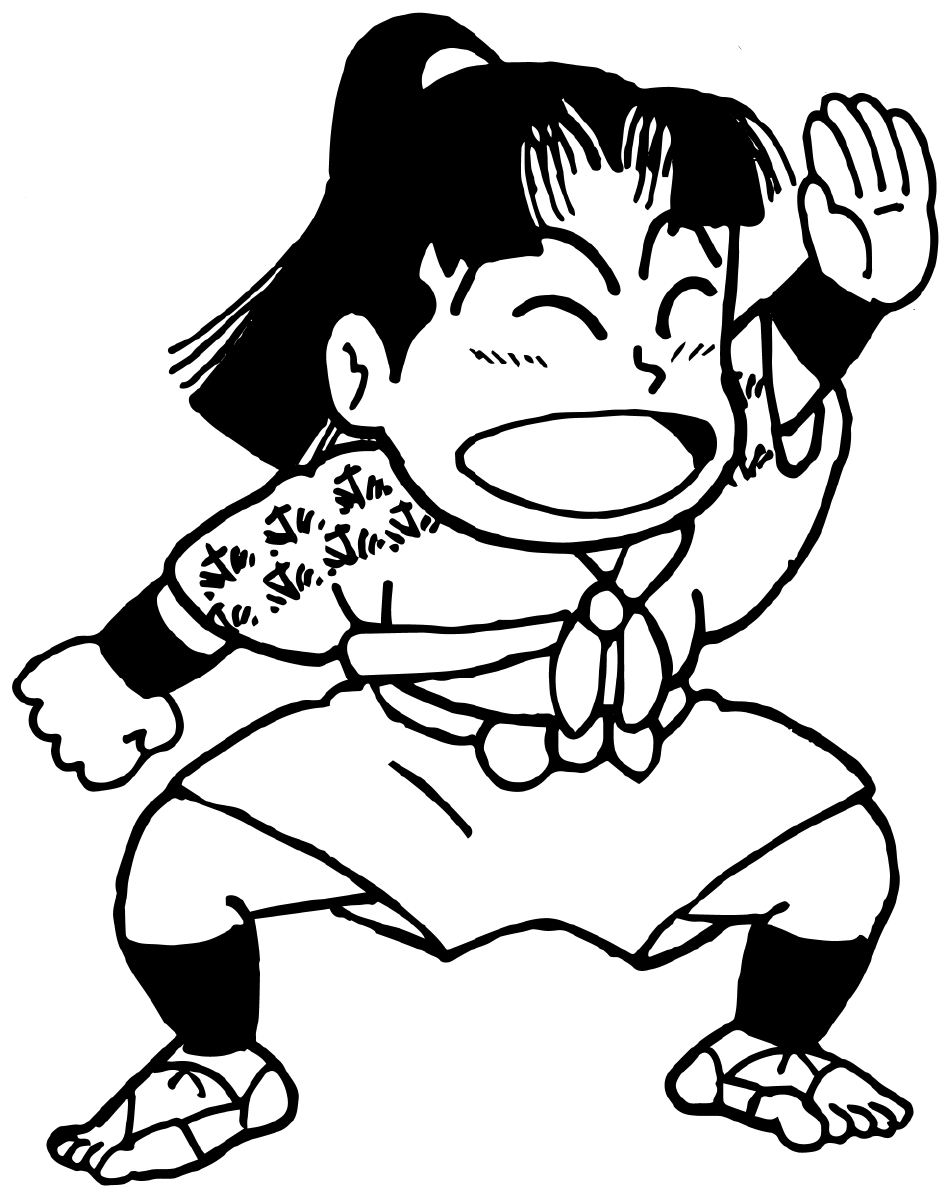}
        }
    \subfloat[screentone-filled]{
        \includegraphics[scale=0.065]{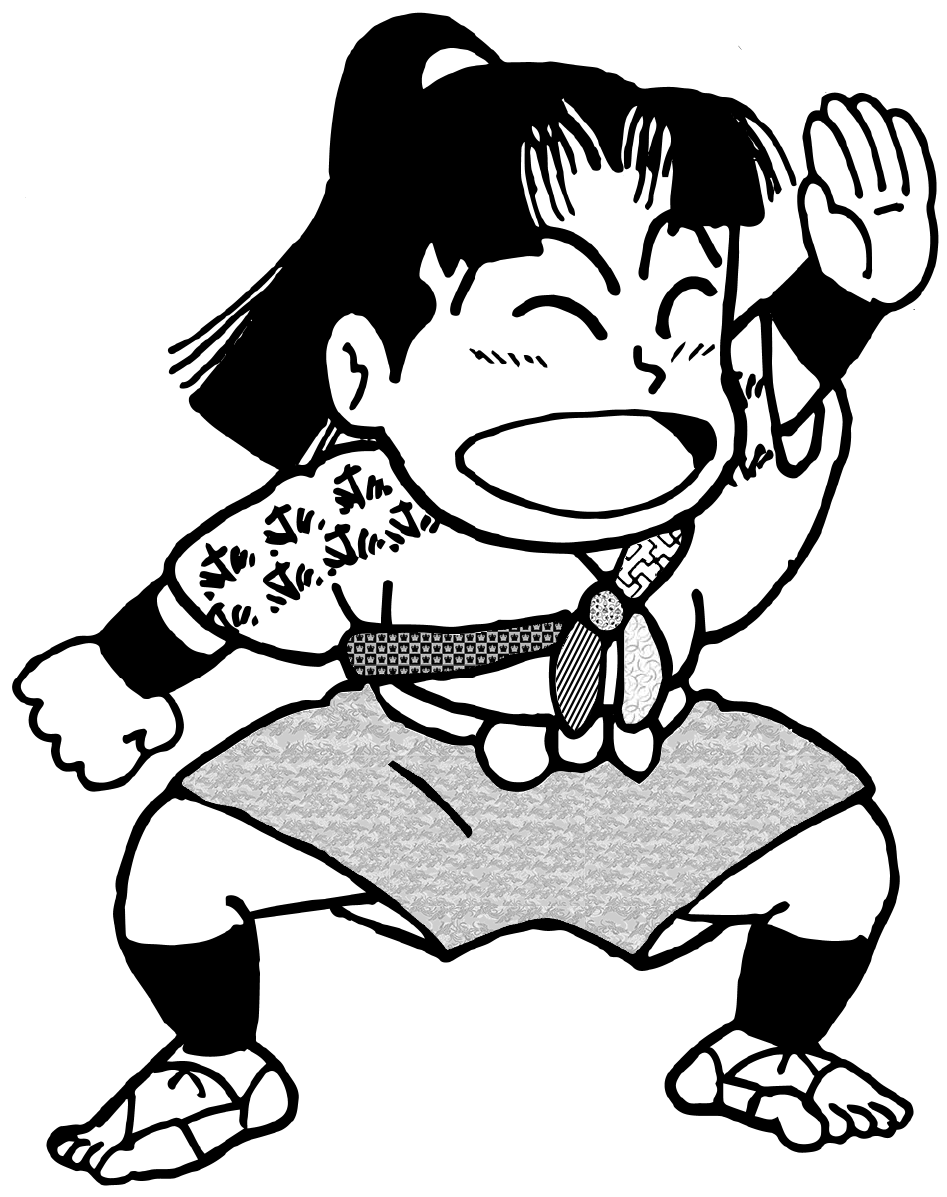}
        }
    \subfloat[mask]{
        \includegraphics[scale=0.065]{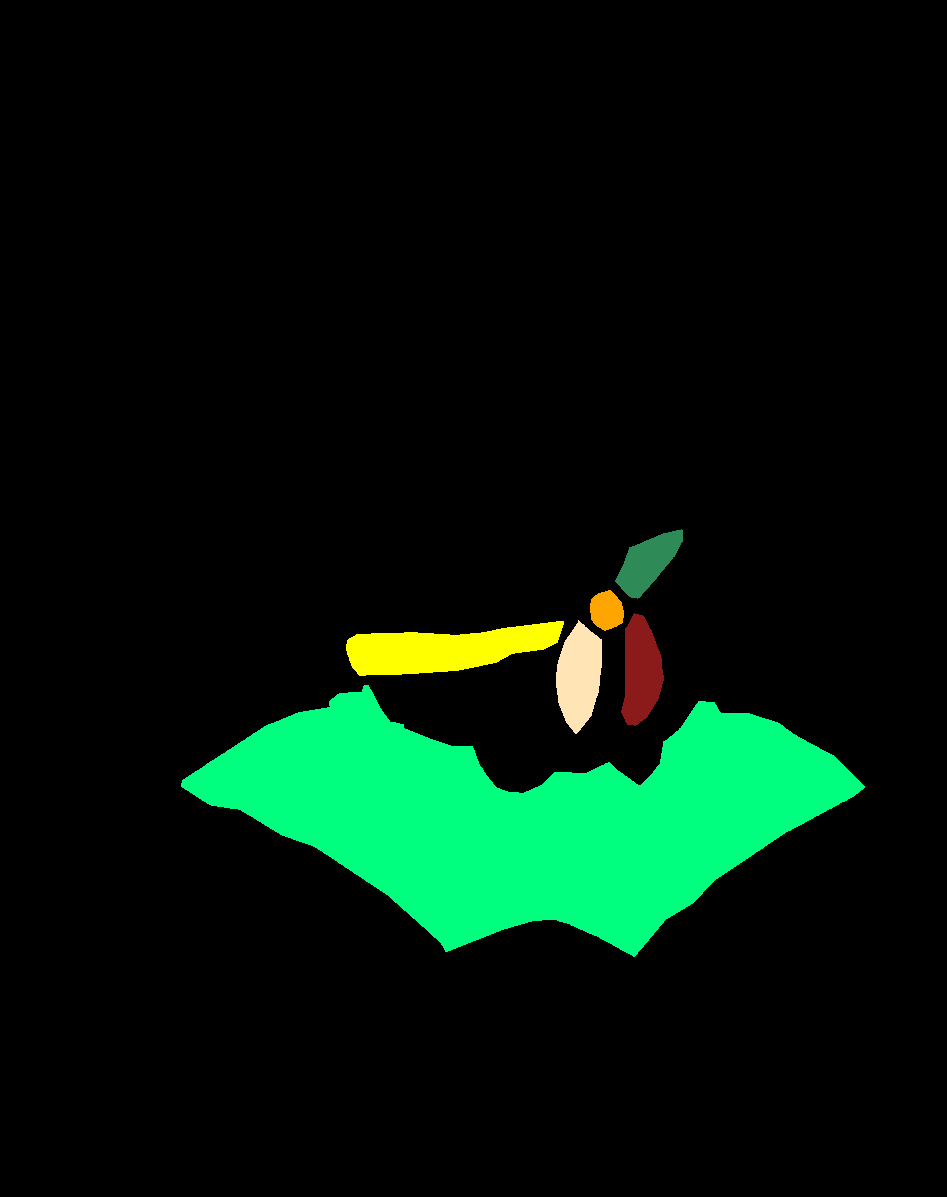}
        }
    \caption{Refill difference type of screentone With the mask to generate training data. }
    \label{CH4_Sec2_fig2}
  \end{center}
\end{figure}

After obtaining line drawings and their corresponding masks, the next step was to fill marked regions on the line drawings with different screentone patterns, and generate masks that were marked with each specific screentone pattern. To accomplish this, we utilized the masks obtained in the previous step. The resulting screentone-filled line drawings and the generated masks are shown in Figure~\ref{CH4_Sec2_fig2}. To visualize the masks, we used different colors to represent different screentone patterns in different regions. Specifically, Figure~\ref{CH4_Sec2_fig2}(a)(d) shows the line drawings, Figure~\ref{CH4_Sec2_fig2}(b)(e) shows the image filled with screentone, and Figure~\ref{CH4_Sec2_fig2}(c)(f) shows the mask respectively.

During the generation of the training data, we applied random displacement, data augmentation, and image flipping to prevent overfitting. By combining line drawings with different screentone patterns, we obtained approximately 18,000 pieces of training data. We set aside 10\% of the data for the test set and trained our final model for 20-30 epochs. This model was then used for semantic segmentation to predict the type and location of screentone in images.

In our study, we utilized the Dice coefficient to evaluate the segmentation results and the Dice loss as the loss function. The Dice coefficient is a widely used statistic for measuring the similarity between two sets. In our case, an image’s number of channels is the number of screentone patterns it has, calculated using Formula ~\ref{CH4_Sec2_eq1} – adding up the Dice coefficients obtained from all channels and then dividing it by the total number of channels. Dice loss can be obtained using 1-Dice coefficient. We calculated the Intersection over Union (IoU) scores for the training and test sets by applying our model. The highest IoU scores we achieved were 0.995 for the training set and 0.978 for the test set.

\begin{equation}
    \frac{1}{c}\sum_{c}{Dice\ coefficient}
    \label{CH4_Sec2_eq1}
\end{equation}

\begin{figure}[htbp]
  \begin{center}
    \includegraphics[scale=0.14]{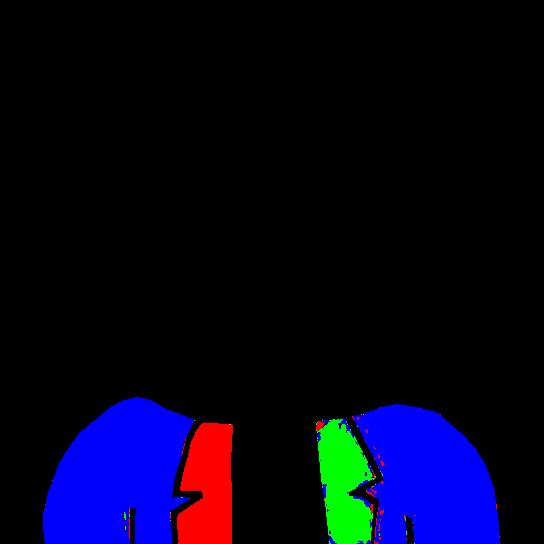}
    \includegraphics[scale=0.14]{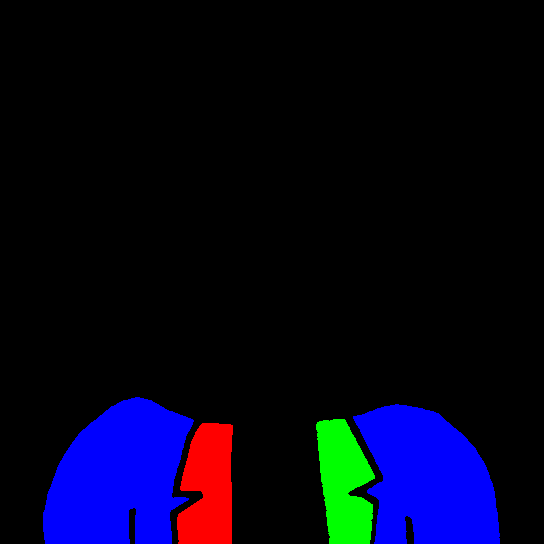}
    \\
    \includegraphics[scale=0.10]{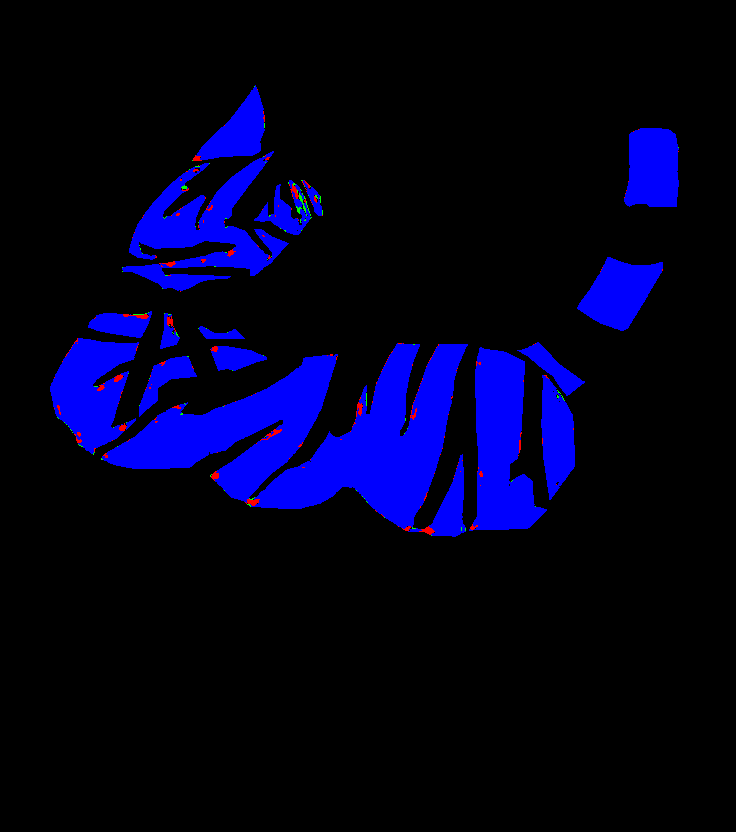}
    \includegraphics[scale=0.10]{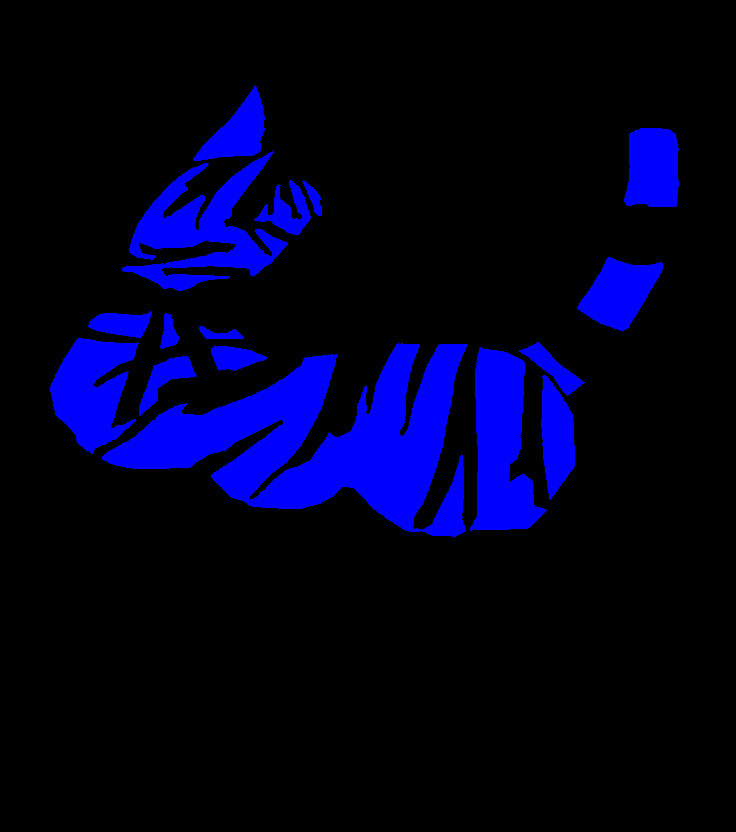}
    \caption{Results from before and after our correction. From left to right, before correction, after correction.}
    \label{CH4_Sec2_fig3}
  \end{center}
\end{figure}

\begin{figure}[htbp]
  \begin{center}
    \includegraphics[scale=0.18]{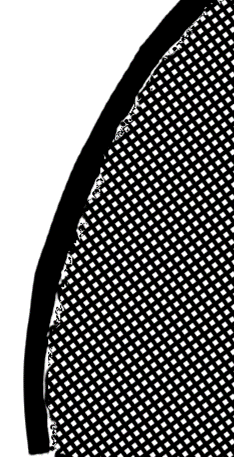}
    \includegraphics[scale=0.18]{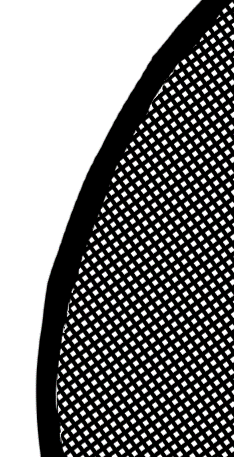}
    \includegraphics[scale=0.04]{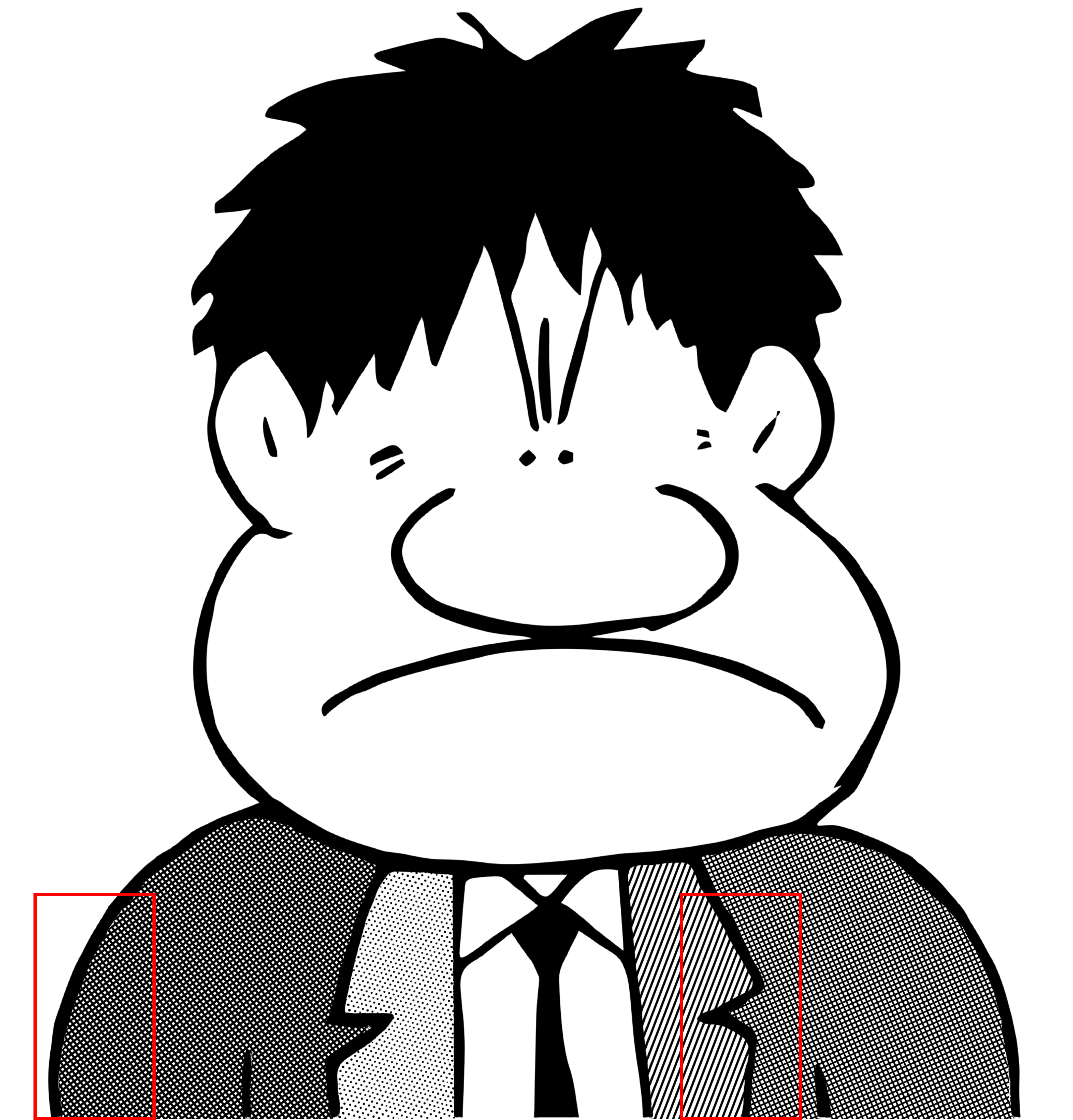}
    \includegraphics[scale=0.18]{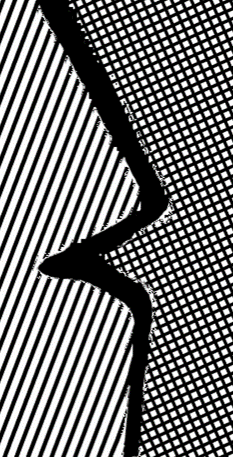}
    \includegraphics[scale=0.18]{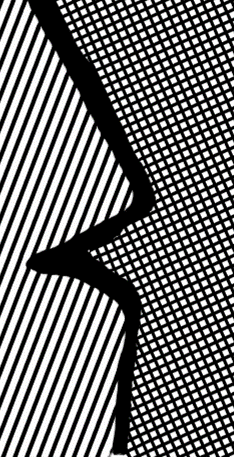}
    \caption{Results from before and after our correction. From left to right, before correction, after correction.}
    \label{CH4_Sec2_fig4}
  \end{center}
\end{figure}

We discovered that the model has a tendency of misjudging detected screentone pattern when it comes to intersecting areas between a screentone region and the comic’s main structural lines, resulting in a noise-like fragmented region. To resolve this problem, we first binarized screentone regions and performed connected component analysis on them because we noticed that regions with different-patterned screentone are often separated by the comic’s main structural lines; in other words, screentone within the same connected region are of the same type. After running connected component analysis, we label a connected component with the screentone pattern that appears most frequently in that region to represent the screentone pattern of that specific connected component; by doing so, we eliminate regions that are given misjudged determination. In addition, we also connected components that are too small because these regions are often products of misdetection. Figure ~\ref{CH4_Sec2_fig3} and ~\ref{CH4_Sec2_fig4} showcases the results from before and after our correction.

\subsection{Super Resolution}
To perform super resolution on comics, this study referenced the RRDB architecture in ESRGAN ~\cite{ESRGAN} in our image restoration of comic screentone. Our goal is to run LR images through super resolution processes to restore them into HR images that at the same time still have the same screentone density and pattern direction as the original LR images, instead of simply magnifying the screentone, so that the HR images still bear the visual effect of how readers should perceive the screentone.

\begin{figure}[htbp]
    \centerline{
        \includegraphics[scale=0.4]{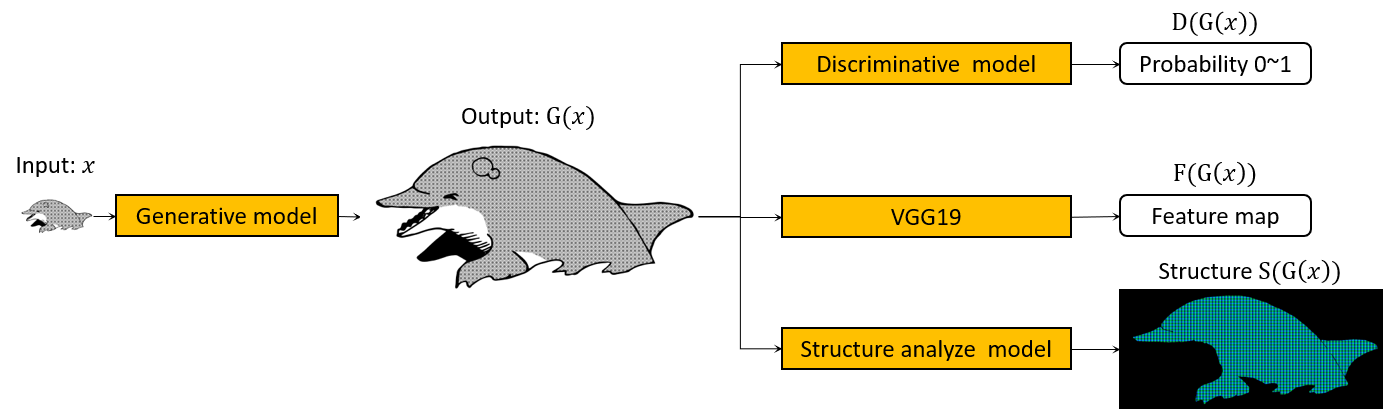}
        }
    \caption{Super Resolution Network Structure}
    \label{CH4_Sec2_fig6}
\end{figure}

Figure~\ref{CH4_Sec2_fig6} offers a comprehensive illustration of our super resolution flowchart. The input image is represented in x. First, the generative model G(x) generate the HR image from the corresponding LR image. Then, the discriminative model D(x) determines whether the image was produced by a generative model or already exists in the training dataset. Different from ESRGAN, we add a structural network for discriminating screentone structure in order to remove the artifacts created by the generative model.

\begin{figure}
  \begin{center}    
    \includegraphics[width=.3\linewidth]{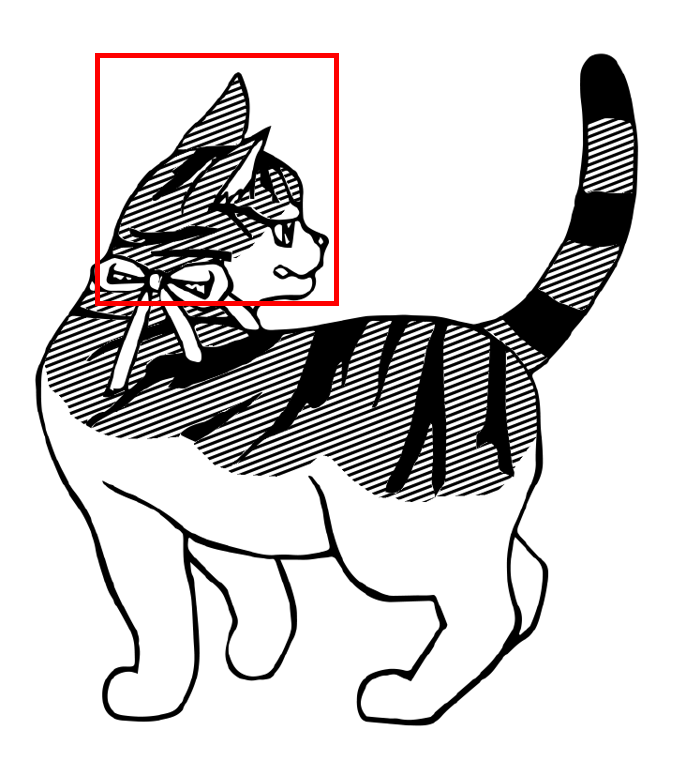}     
    \includegraphics[width=.3\linewidth]{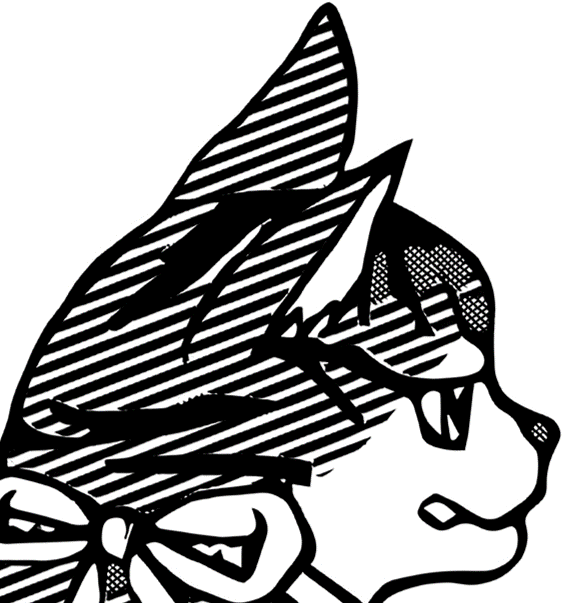}
    \includegraphics[width=.3\linewidth]{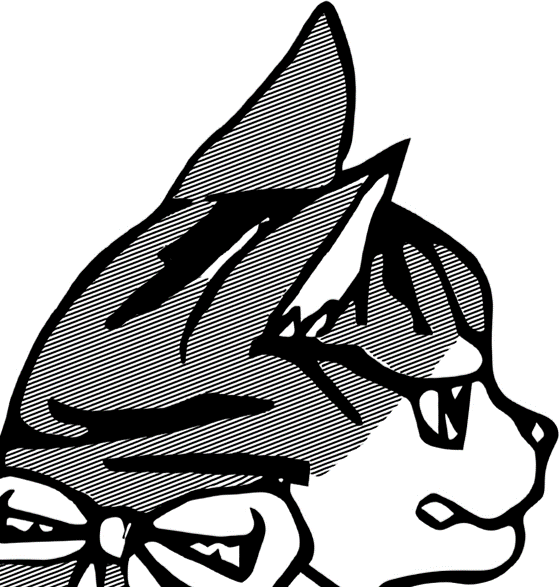}
    \\
    \includegraphics[width=.2\linewidth]{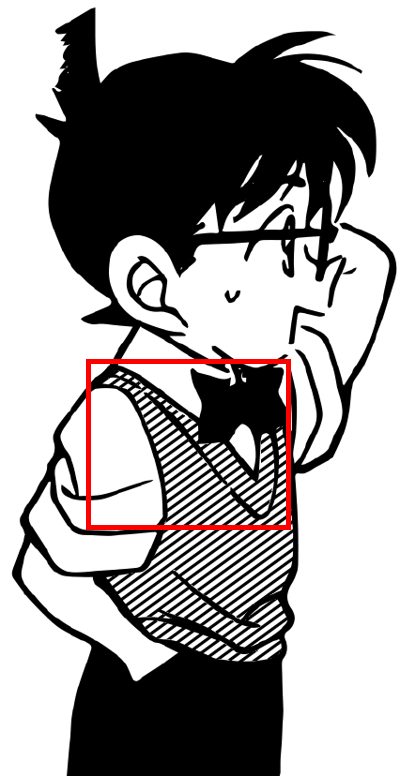}
    \includegraphics[width=.3\linewidth]{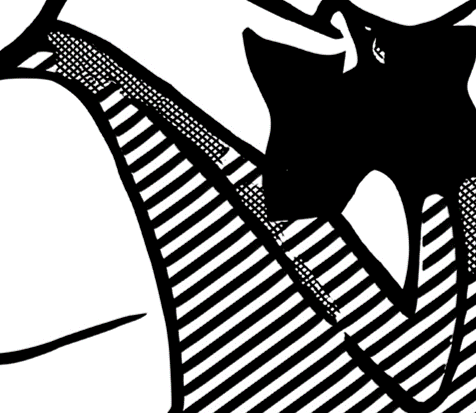}
    \includegraphics[width=.3\linewidth]{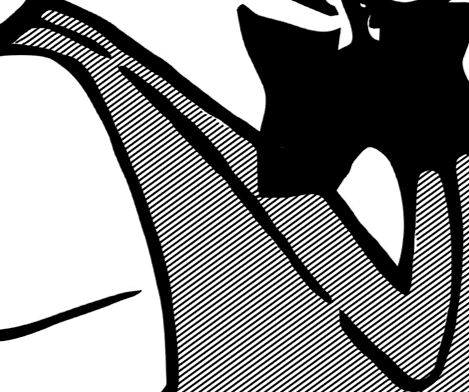}
    \\    
    \includegraphics[width=.3\linewidth]{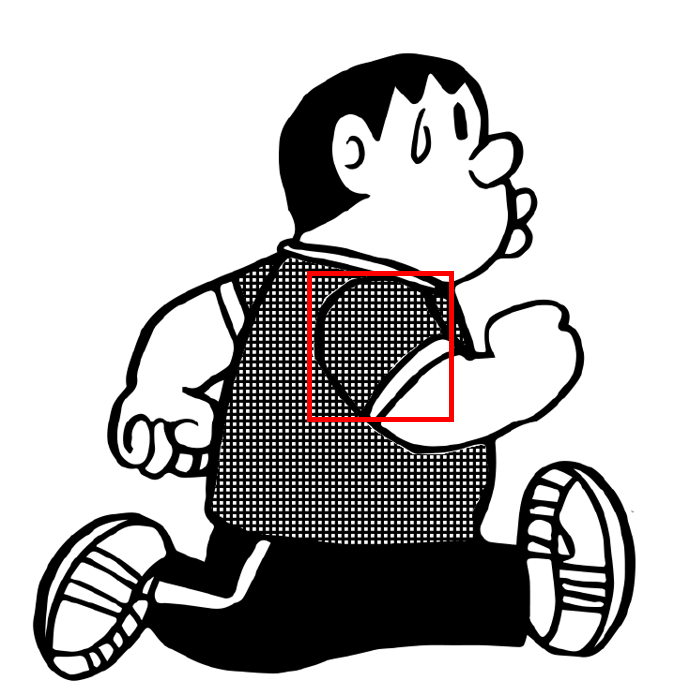}       
    \includegraphics[width=.3\linewidth]{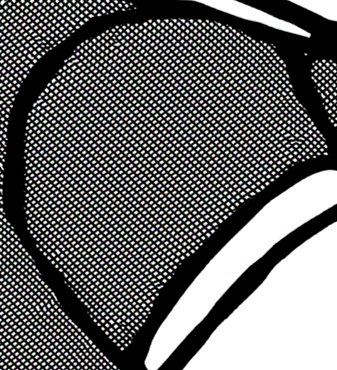}       
    \includegraphics[width=.3\linewidth]{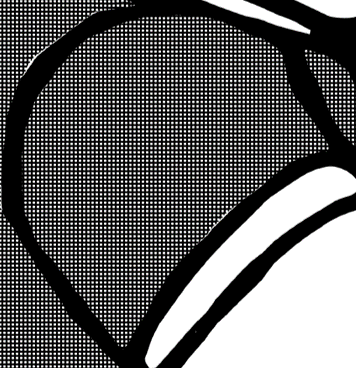}
    \\
    \subfloat[LR]{
        \includegraphics[width=.3\linewidth]{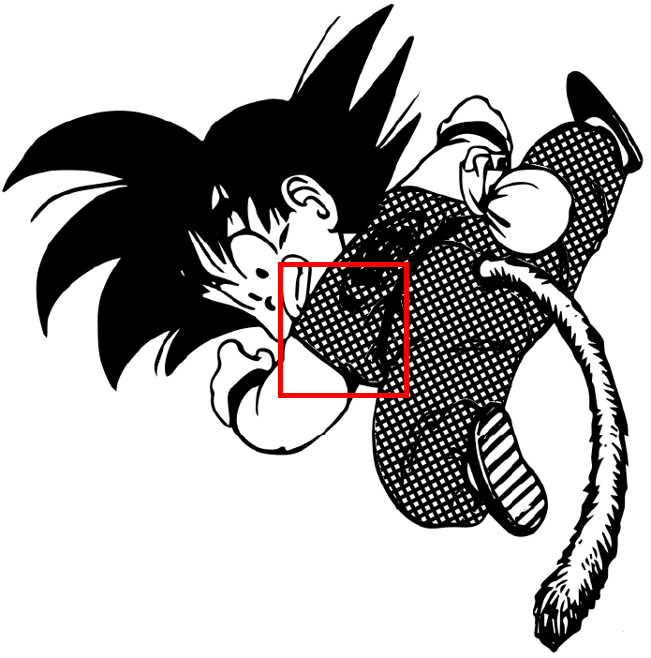}
        }
    \subfloat[Mixed Training]{
        \includegraphics[width=.3\linewidth]{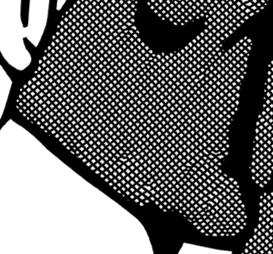}
        }
    \subfloat[Separate Training]{
        \includegraphics[width=.3\linewidth]{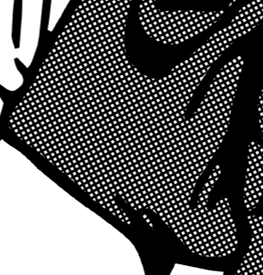}
        }
  \end{center}
   \caption{Compare the super-resolution results obtained by models trained with mixed training using different datasets and models trained separately.}
    \label{CH4_Sec2_fig7}
\end{figure}

We first tried filling line drawings with three types of screentone – dotted, linear, gridded – for test training, with results detailed in Figure \ref{CH4_Sec2_fig7}. In Figure \ref{CH4_Sec2_fig7}(a), the LR image on the left undergoes super resolution using the trained model with results shown on the right. We can see that the model misidentified the crossing between linear screentone and the manga's main structural lines as gridded screentone and therefore restoring that region into grids. Meanwhile, Figure \ref{CH4_Sec2_fig7}(b) demonstrates how the gridded screentone displayed distorted lines after image restoration, resulting in bad effects.
Through the above experiment, we came to the conclusion that a super resolution model trained with mixed types of screentone cannot fulfill our expectation; hence, we decided to first apply image semantic segmentation to classify screentone in LR images and then conduct our super resolution process using different models that matched specific screentone patterns. 

\begin{figure}[htbp]
  \begin{center}
    \subfloat[LR]{
        \includegraphics[scale=0.05]{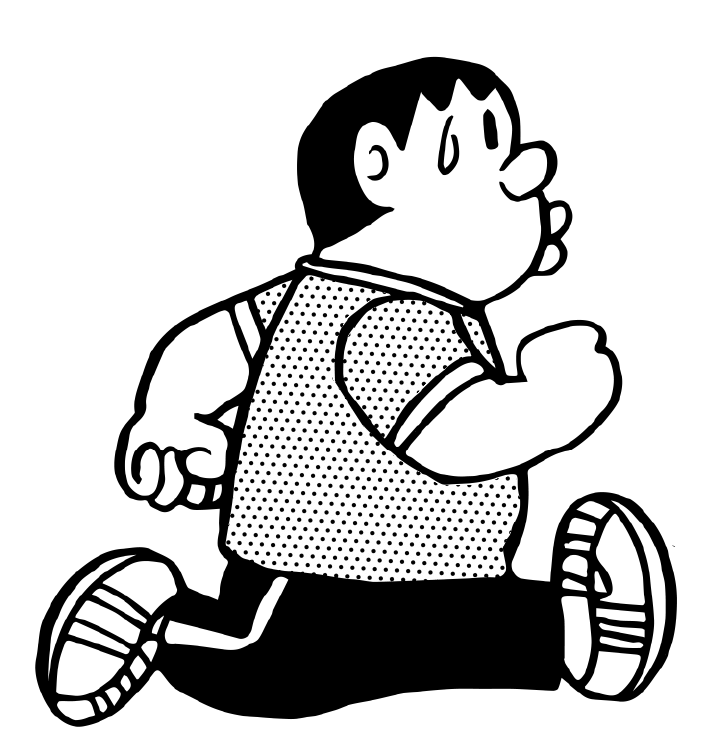}
        }
    \subfloat[HR]{
        \includegraphics[scale=0.05]{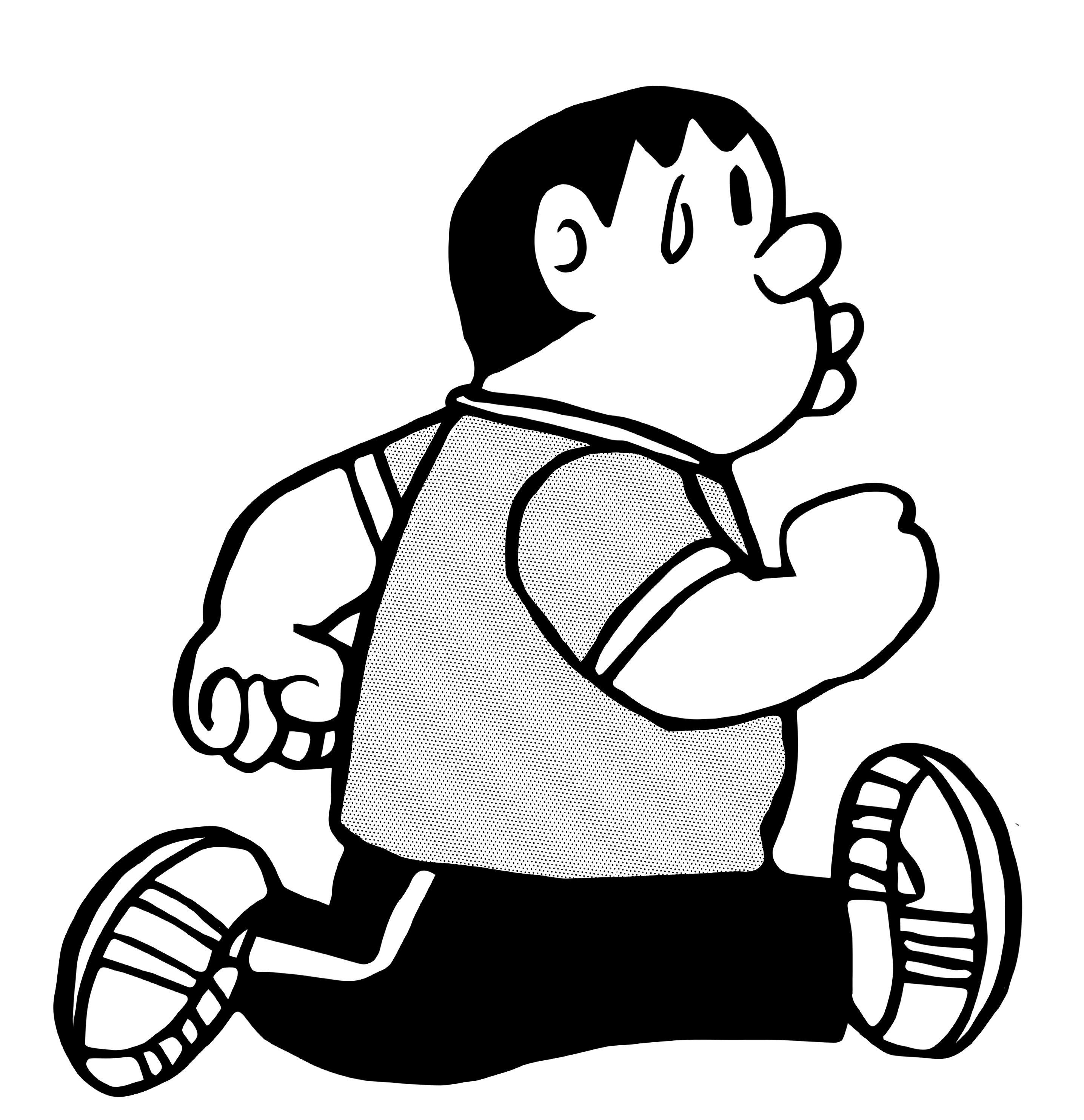}
    }
    \caption{Training Data. We first masked the range of the screentones, and then synthesize different type of screentones to both LR and HR image for more training data. }
    \label{CH4_Sec3_fig1}
  \end{center}
\end{figure}

In terms of training data, we use the same line drawings from semantic segmentation and their corresponding masks, filling both LR and HR line drawings with screentone of same density and pattern direction in order to generate the HR restoration image that we are looking for. Figure~\ref{CH4_Sec3_fig1} shows a set of our training data containing low resolution image(Figure~\ref{CH4_Sec3_fig1}(a)) and high resolution image(Figure~\ref{CH4_Sec3_fig1}(b))e used about 30 pieces of line drawings and screentone with dots, lines, and grids of varying thickness, density and direction; for each type of screentone pattern, we generated around 6000 sets of training data that ran for around 500 epochs before attaining our final model. We utilized this model to perform restoration on LR images; however, it gives rise to several problems as shown in Figure~\ref{CH4_Sec2_fig7} – the system fails to restore the original screentone density because it discriminates linear screentone as the comic’s main structural lines; it rendered  gridded instead of linear screentone due to erroneous discrimination of an intersection between linear screentone and main structural lines; gridded screentone displayed signs of distortion after image restoration.

Given the above reasons, it is our contention that screentone data of different patterns should not be mixed and a super resolution model trained with such data is incapable of achieving the ideal results. Following these, we independently trained super resolution models for different types of screentone pattern. Using the same 30 pieces of line drawings and their corresponding masks, we filled them with different screentone patterns and generated around 6000 sets of training data for each pattern and ran around 500 epochs each. 

\begin{figure}[htbp]
    \centerline{
        \includegraphics[scale=0.3]{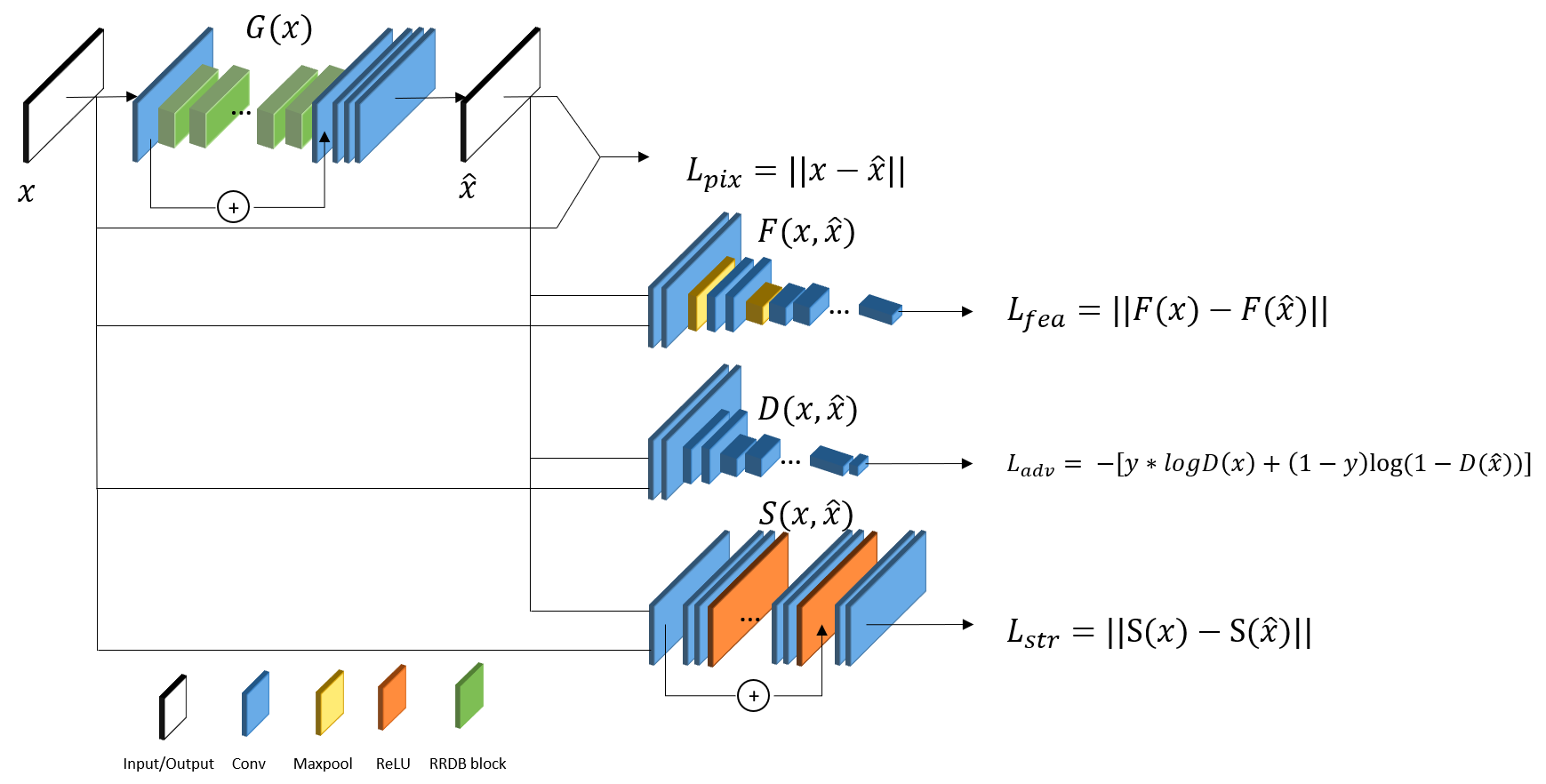}
        }
    \caption{Network Loss Function}
    \label{CH4_Sec3_fig2}
\end{figure}

The loss function is detailed in Figure~\ref{CH4_Sec3_fig2}. At first, we employed three functions – pixel loss $L_{pix}$, feature loss $L_{fea}$, and GAN loss $L_{adv}$. Pixel loss is the calculated L1 loss with truth and restored images as input; feature loss is obtained after running ground truth values and restored images through VGG19 to calculate high-dimensional features and then calculating the L1 loss; GAN loss is an assessment of whether the generated image restoration is detectable to the discriminator.

\begin{figure}[htbp]
  \begin{center}
    \includegraphics[scale=0.25]{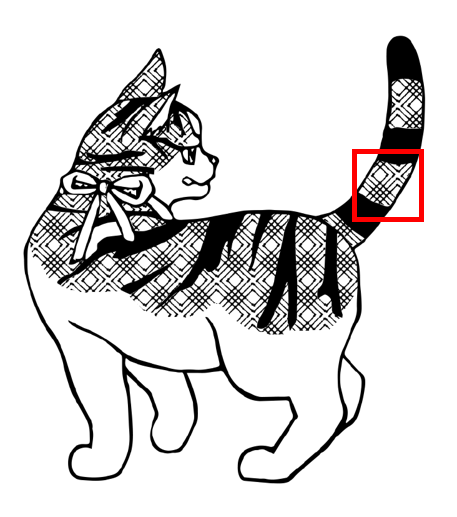}
    \includegraphics[scale=0.5]{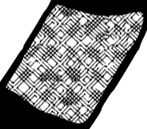}
    \\
    \includegraphics[scale=0.25]{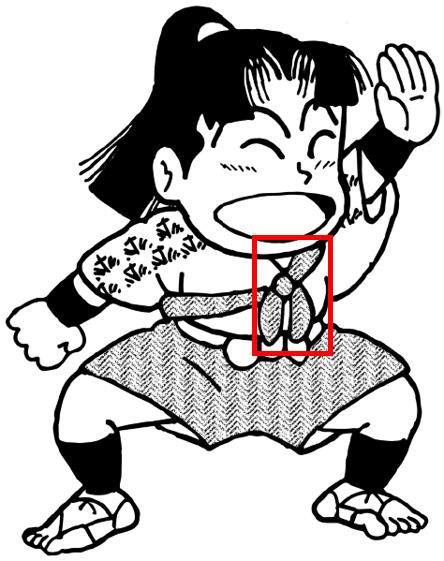}
    \includegraphics[scale=0.35]{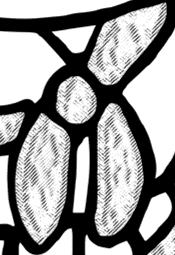}
    \caption{Artifacts created by the model. Image left is LR, right is SR of the region in the red box.}
    \label{CH4_Sec3_fig3}
  \end{center}
\end{figure}

\begin{figure}[htbp]
  \begin{center}
    \includegraphics[scale=0.14]{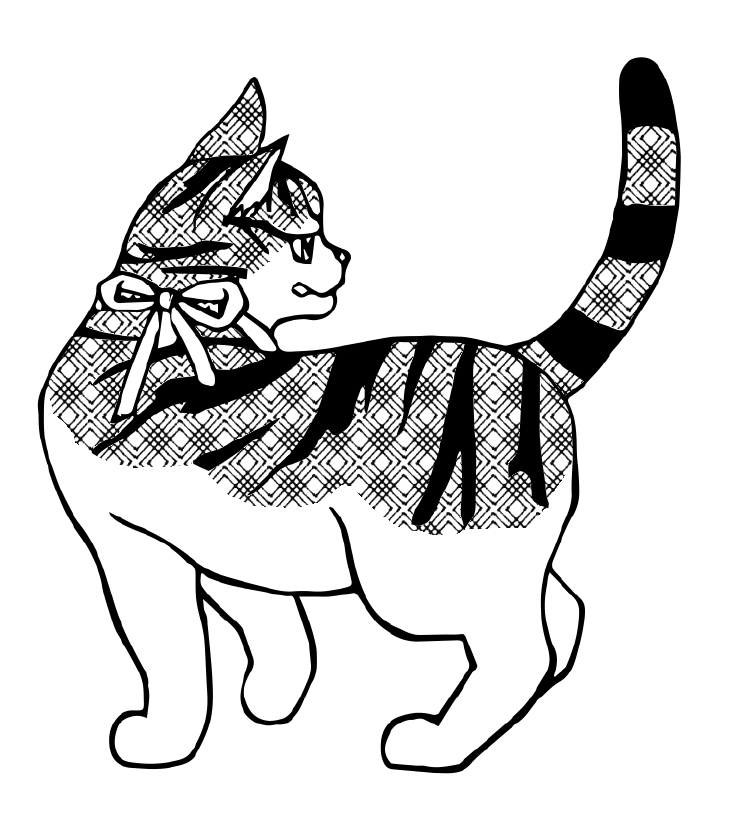}
    \includegraphics[scale=0.14]{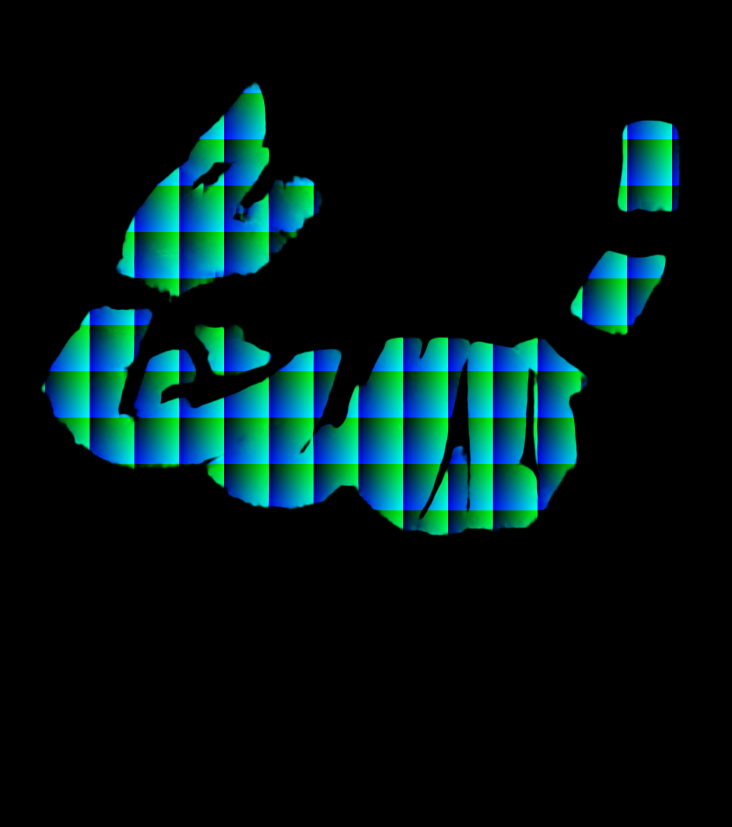}
    \caption{Image and corresponding visualization of its mask.}
    \label{CH4_Sec3_fig4}
  \end{center}
\end{figure}

We found that images restored by models trained with these three loss functions created artifacts in cases involving screentone that had stronger structure, as shown in Figure~\ref{CH4_Sec3_fig3}. We contended that adding structural conditions to loss functions $L_{str}$ would give the model better convergence performance with such types of screentone. We designed an additional network directed for discriminating screentone structure, and then used this network to evaluate structural differences between generated restoration images and ground truth values. In addition to training generator and discriminator, we also trained structural network. For the structural network, the input was images containing screentone while the output was two-dimensional mask images, mapping the offset from each pixel point to the center point of that screentone unit. Figure~\ref{CH4_Sec3_fig4} right is a visualization of these masks.

\begin{figure}[htbp]
  \begin{center}
    \includegraphics[scale=0.25]{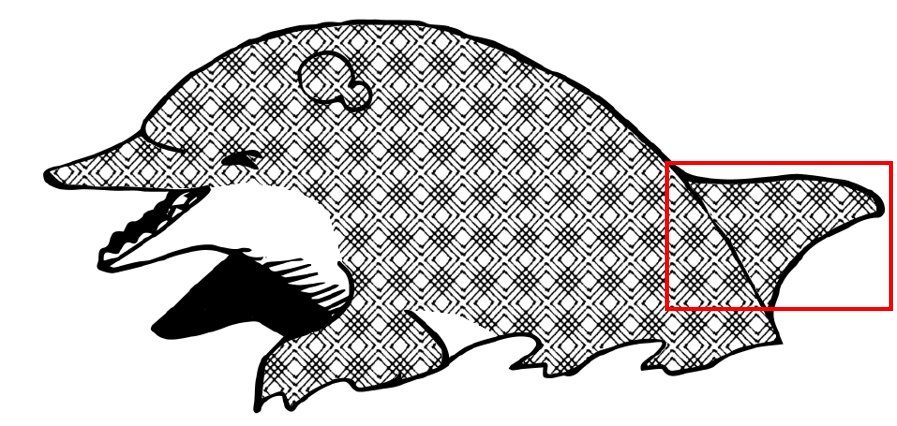}
    \includegraphics[scale=0.15]{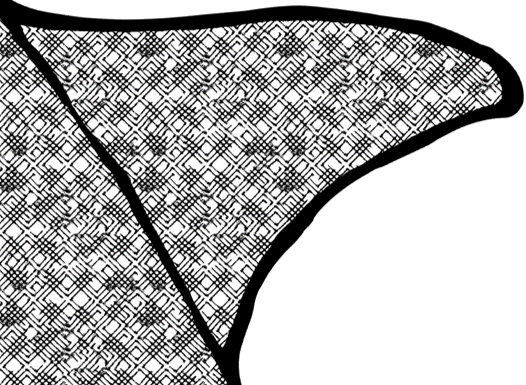}
    \includegraphics[scale=0.15]{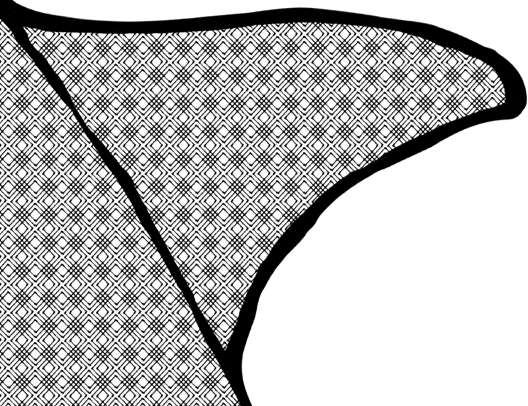}
    \\
    \includegraphics[scale=0.2]{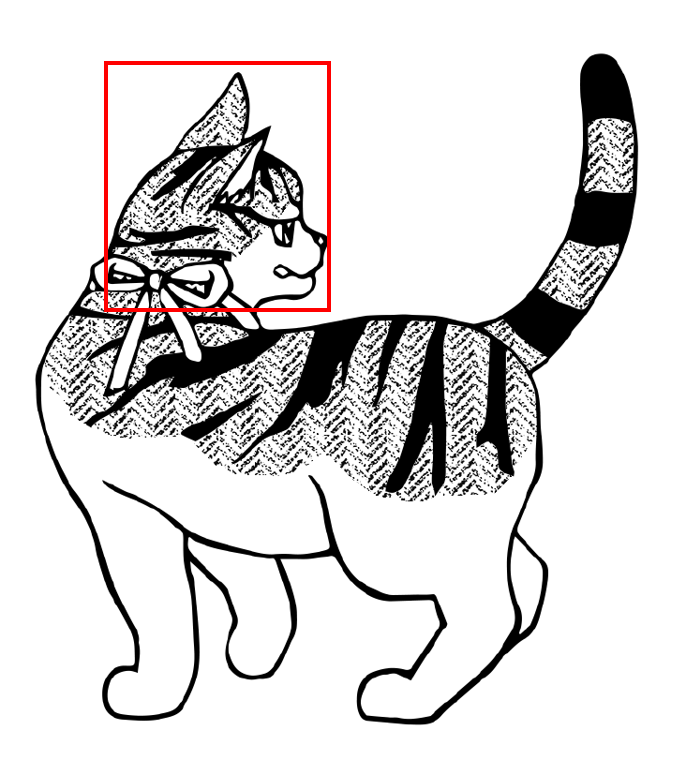}
    \includegraphics[scale=0.15]{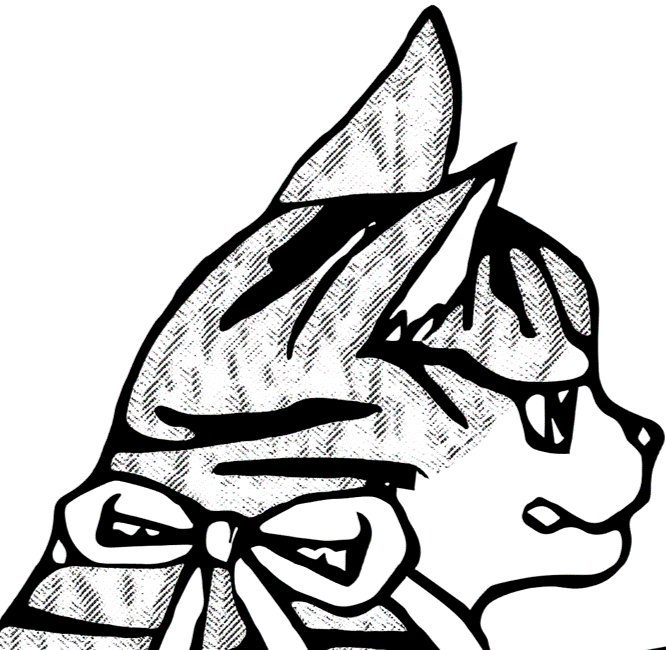}
    \includegraphics[scale=0.15]{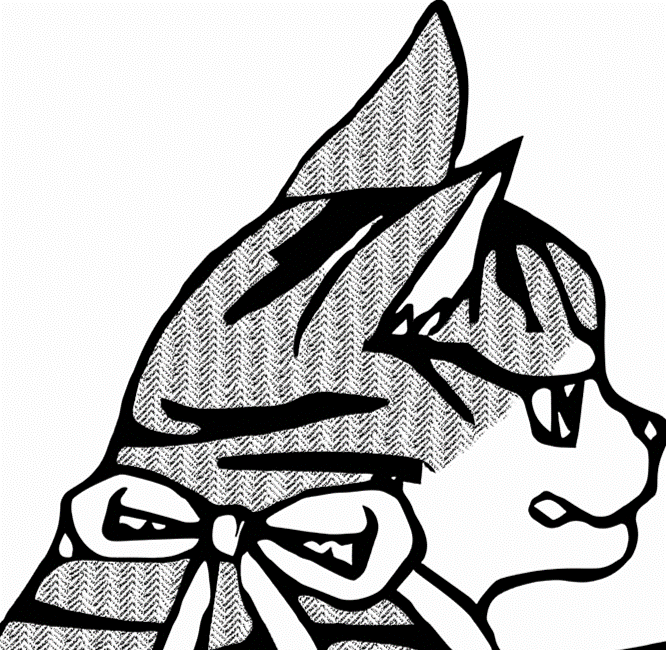}
    \\
    \includegraphics[scale=0.25]{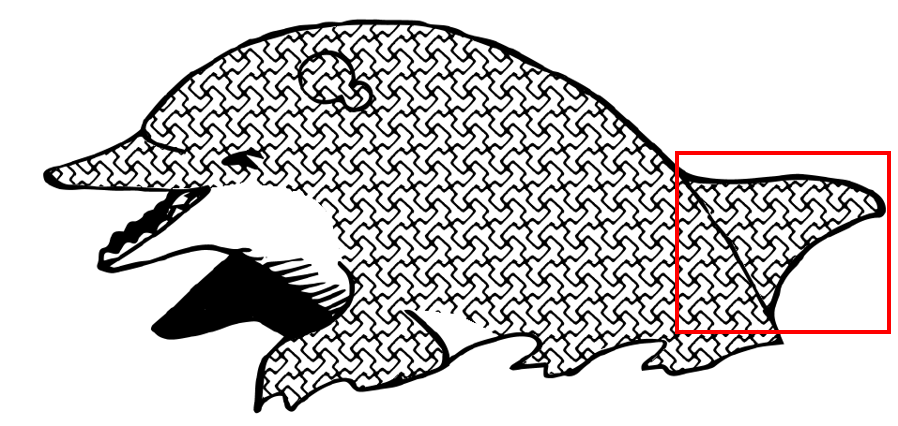}
    \includegraphics[scale=0.15]{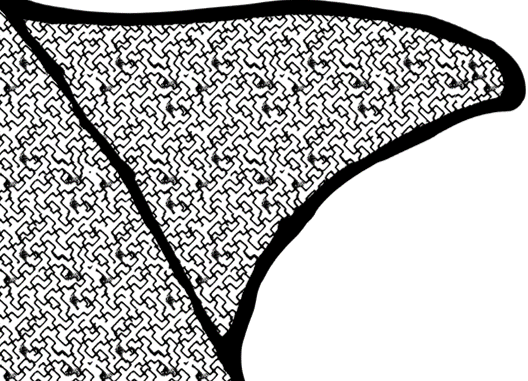}
    \includegraphics[scale=0.15]{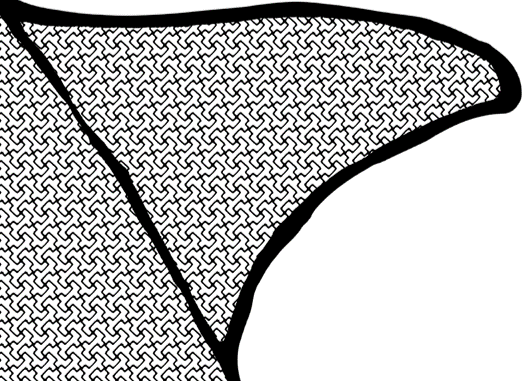}
    \\
    \includegraphics[scale=0.25]{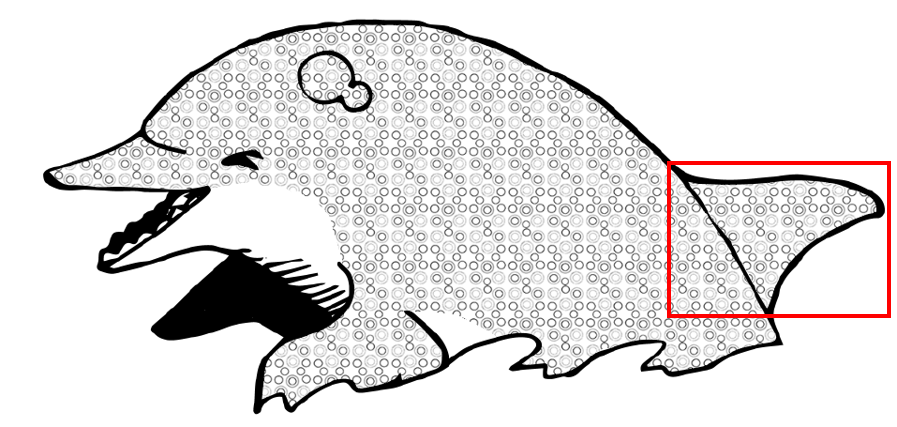}
    \includegraphics[scale=0.15]{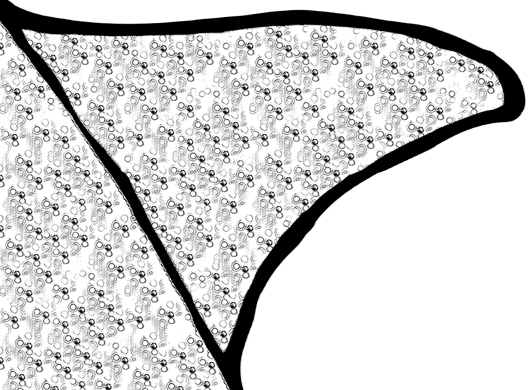}
    \includegraphics[scale=0.15]{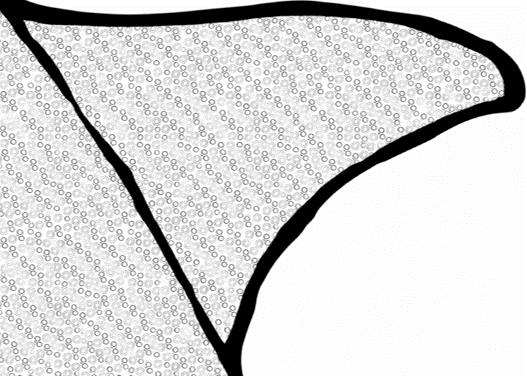}
    \caption{LR image, without structural network and with structural network.}
    \label{CH4_Sec3_fig5}
  \end{center}
\end{figure}

Figure~\ref{CH4_Sec3_fig5} depicts a result comparison of and without structural network adding structural network. Adding structural network is able to resolve the artifacts issue.

Finally, we can enlarge the low-resolution manga and retain the density of screentone using super resolution method. We take take advantage of generative adversarial network to preserve more detail and keep the density of screentone instead of simply interpolate to magnify the images. Moreover, we can keep the structure of the screentone and prevent the artifacts effectively by adding the structural constraint into our network.

\section{Experiment Retults and Discussion}
\subsection{Experiment Retults}
\begin{figure}[htbp]
  \begin{center}
    \subfloat[Size:3, Gap:3, Angle:30]{
        \includegraphics[scale=0.2]{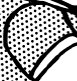}
        \includegraphics[scale=0.2]{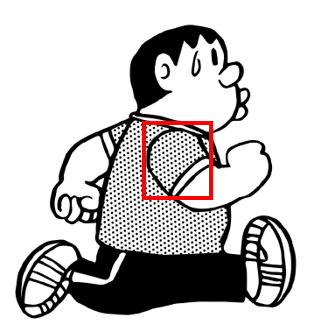}
        \includegraphics[scale=0.2]{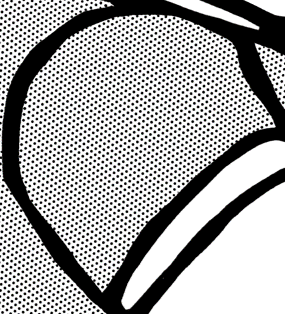}
    }
    \subfloat[Width:4, Gap:4, Angle:83]{
        \includegraphics[scale=0.22]{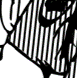}
        \includegraphics[scale=0.22]{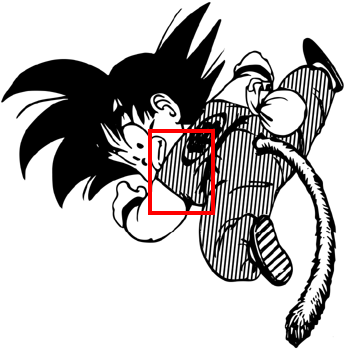}
        \includegraphics[scale=0.22]{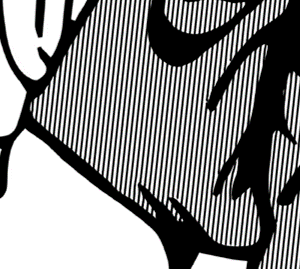}
    }
    \\
    \subfloat[Width:3, Gap:6, Angle:71]{
        \includegraphics[scale=0.16]{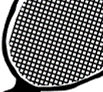}
        \hspace{0.6mm}
        \includegraphics[scale=0.16]{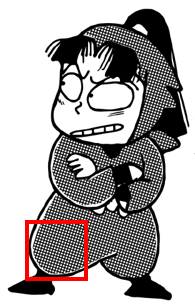}
        \hspace{0.6mm}
        \includegraphics[scale=0.16]{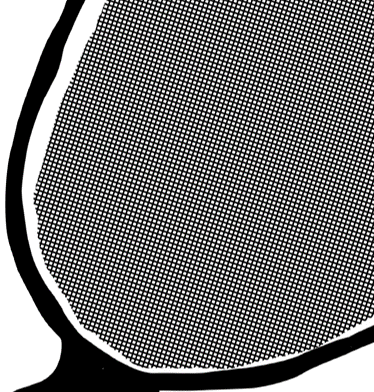}
    }
    \subfloat[Width:3, Gap:4, Angle:21]{
        \includegraphics[scale=0.25]{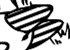}
        \includegraphics[scale=0.25]{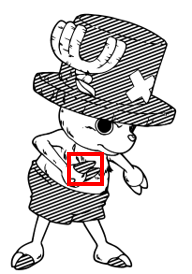}
        \includegraphics[scale=0.25]{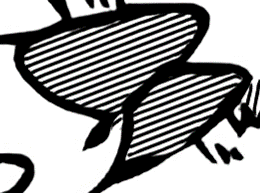}
    }
    \caption{Super resolution results about dotted, linear, and gridded screentone. The last on the left is the area in the red box of LR image, and the last on the right is the area in the red box of SR image.}
    \label{CH5_Sec1_fig1}
  \end{center}
\end{figure}

\begin{figure}[htbp]
    \centering
    \subfloat[Size:11]{
        \includegraphics[scale=0.25]{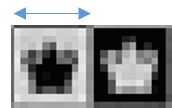}
        \includegraphics[scale=0.08]{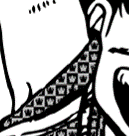}
        \includegraphics[scale=0.08]{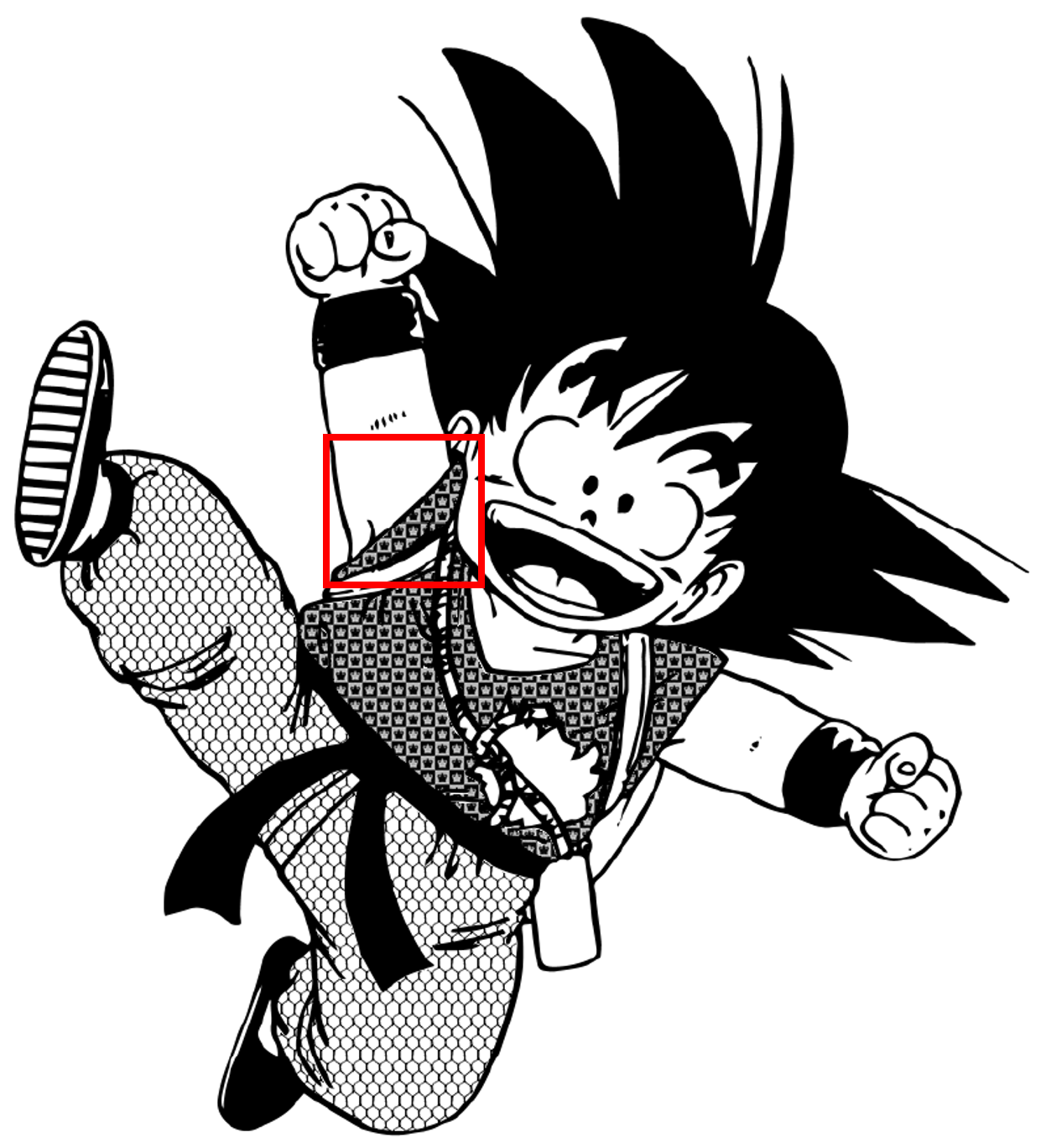}
        \includegraphics[scale=0.08]{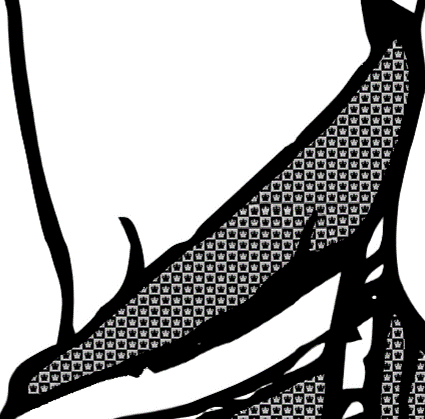}
    }
    \subfloat[Size:9]{
        \includegraphics[scale=0.25]{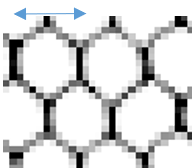}
        \includegraphics[scale=0.08]{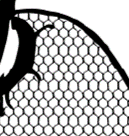}
        \includegraphics[scale=0.08]{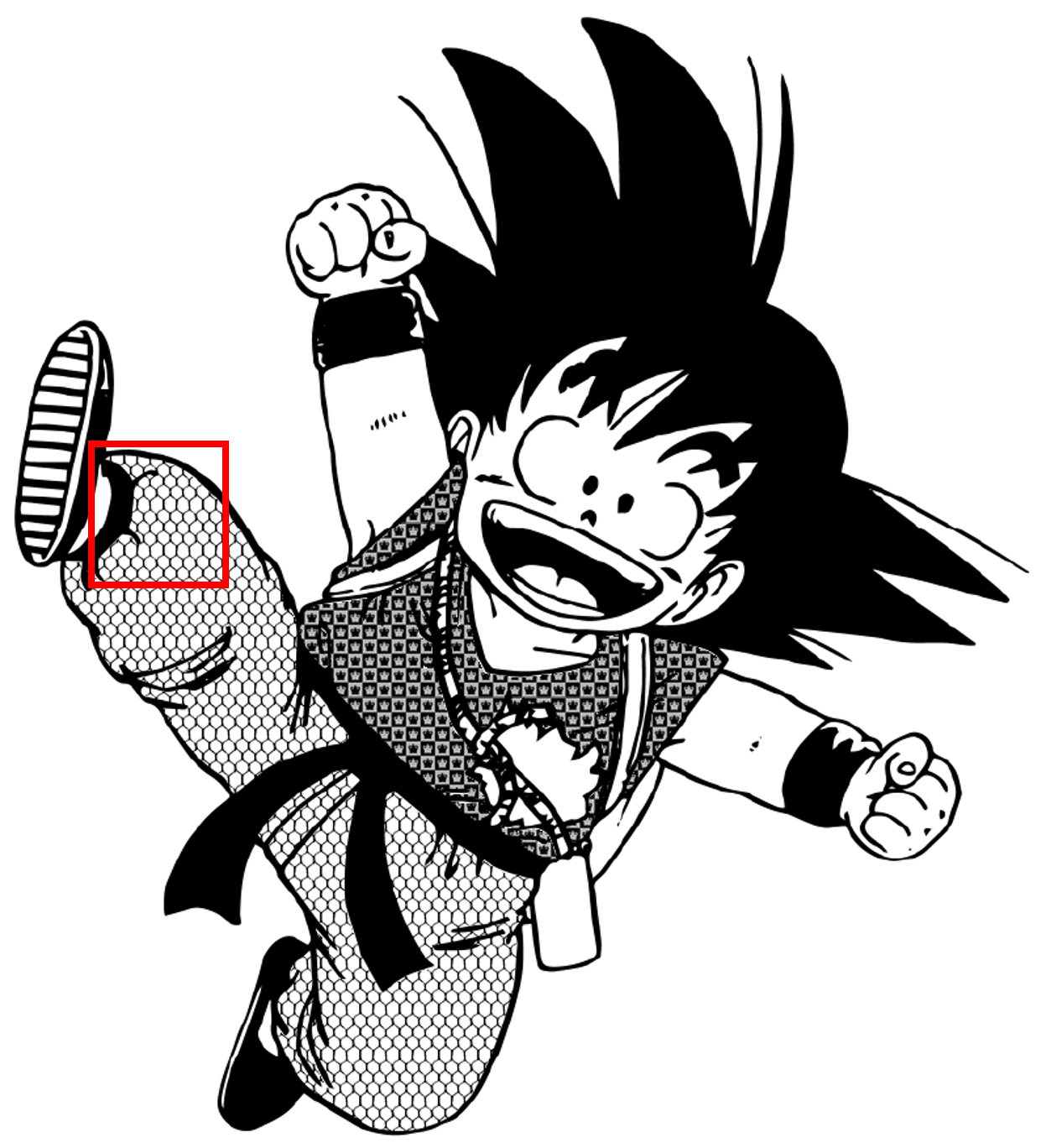}
        \includegraphics[scale=0.08]{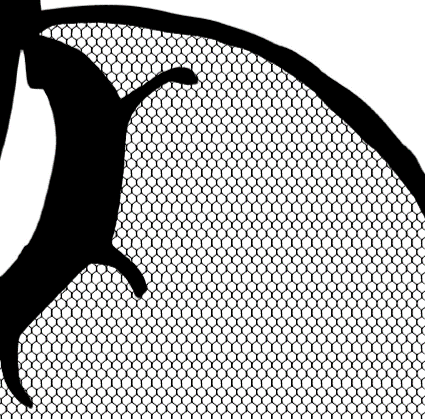}
    }
    \\
    \subfloat[Size:44]{
        \includegraphics[scale=0.3]{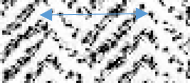}
        \includegraphics[scale=0.12]{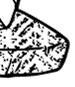}
        \includegraphics[scale=0.12]{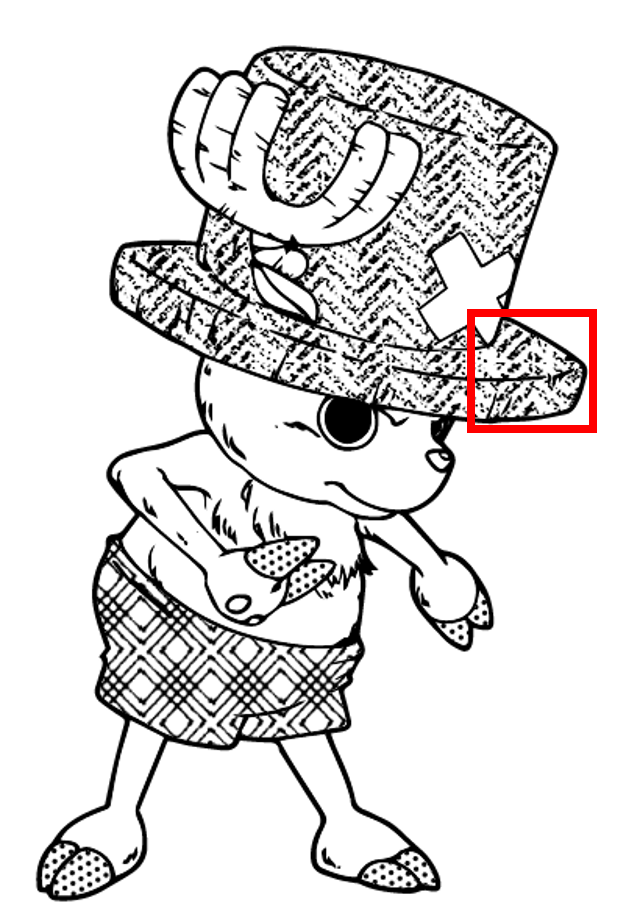}
        \includegraphics[scale=0.12]{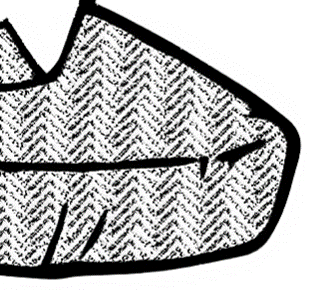}
    }
    \subfloat[Size:43]{
        \includegraphics[scale=0.3]{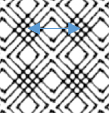}
        \includegraphics[scale=0.12]{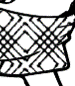}
        \includegraphics[scale=0.12]{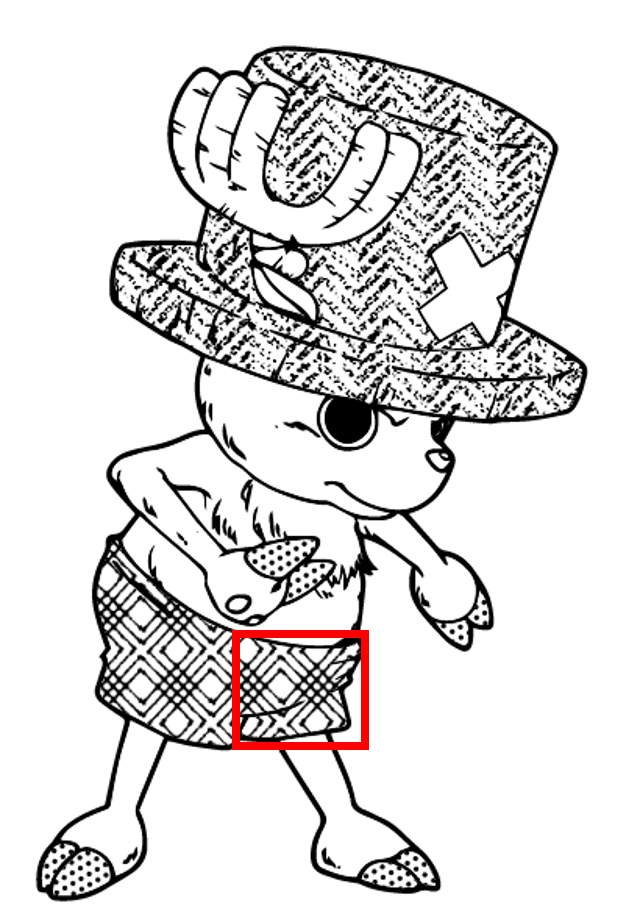}
        \includegraphics[scale=0.12]{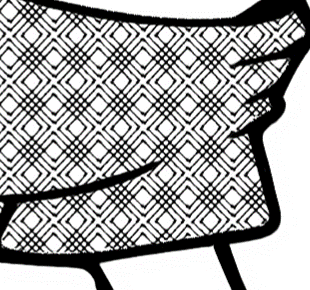}
    }
    \\
    \subfloat[Size:11]{
        \includegraphics[scale=0.3]{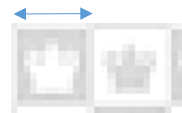}
        \includegraphics[scale=0.10]{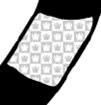}
        \includegraphics[scale=0.10]{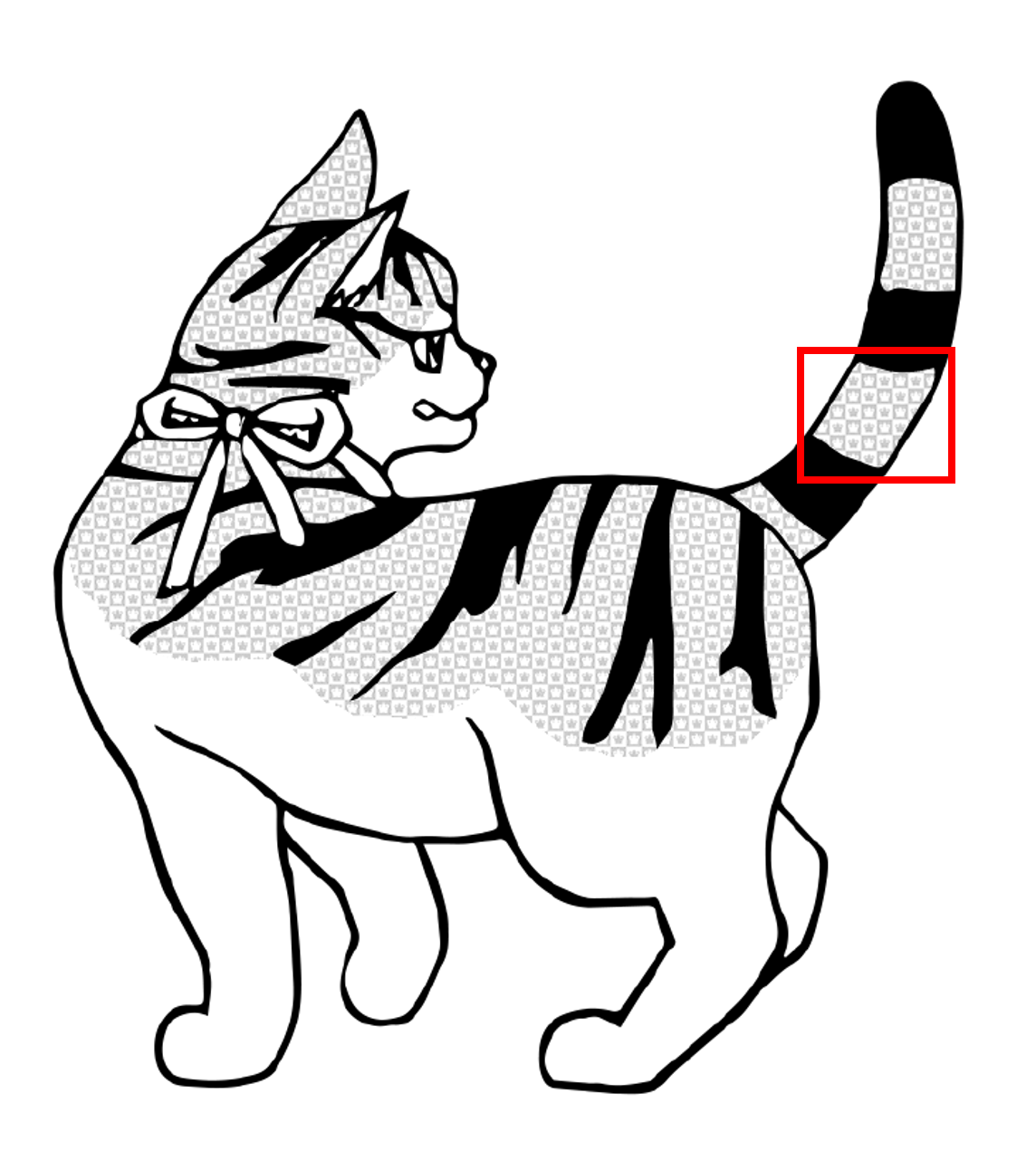}
        \includegraphics[scale=0.10]{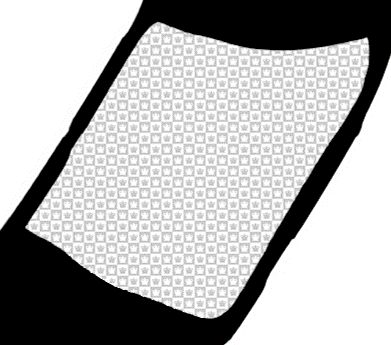}
    }
    \subfloat[Size:41]{
        \includegraphics[scale=0.3]{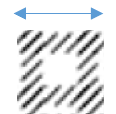}
        \includegraphics[scale=0.08]{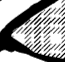}
        \includegraphics[scale=0.08]{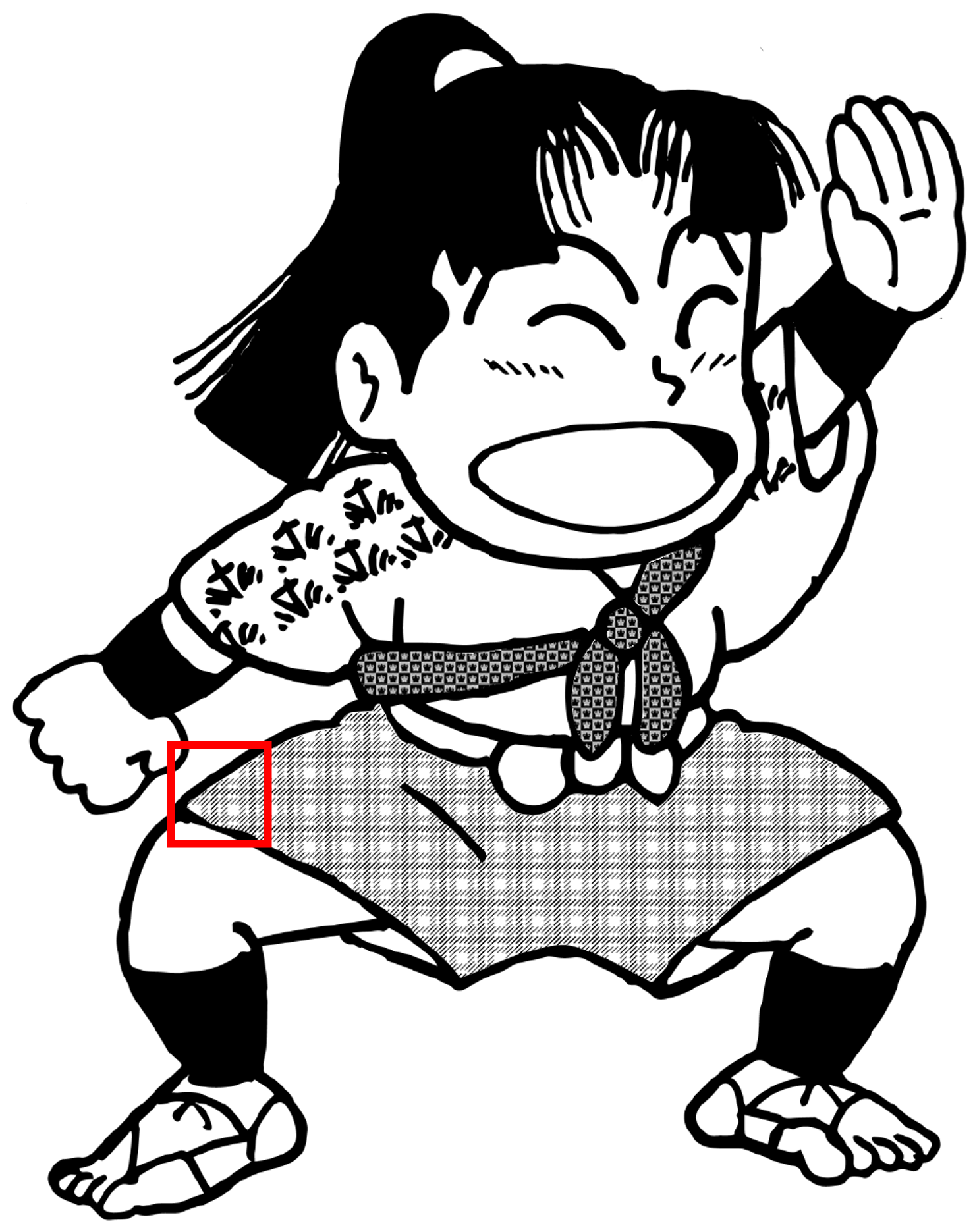}
        \includegraphics[scale=0.08]{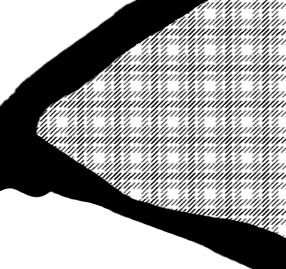}
    }
    \\
    \subfloat[Size:16]{
        \includegraphics[scale=0.3]{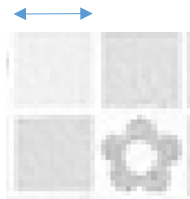}
        \includegraphics[scale=0.08]{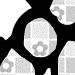}
        \includegraphics[scale=0.08]{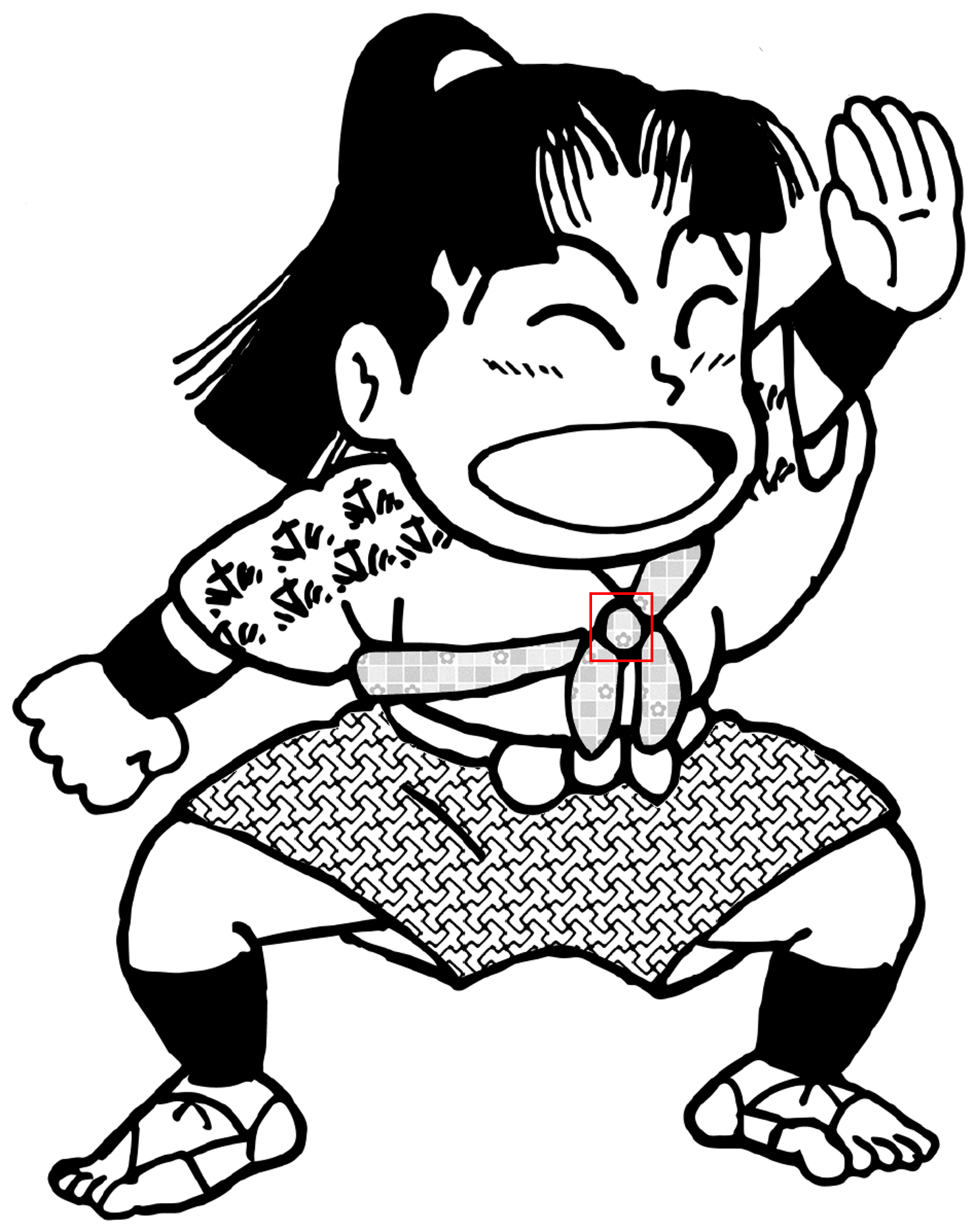}
        \includegraphics[scale=0.08]{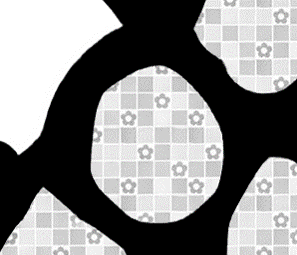}
    }
    \subfloat[Size:40]{
        \includegraphics[scale=0.3]{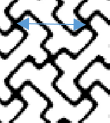}
        \includegraphics[scale=0.08]{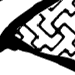}
        \includegraphics[scale=0.08]{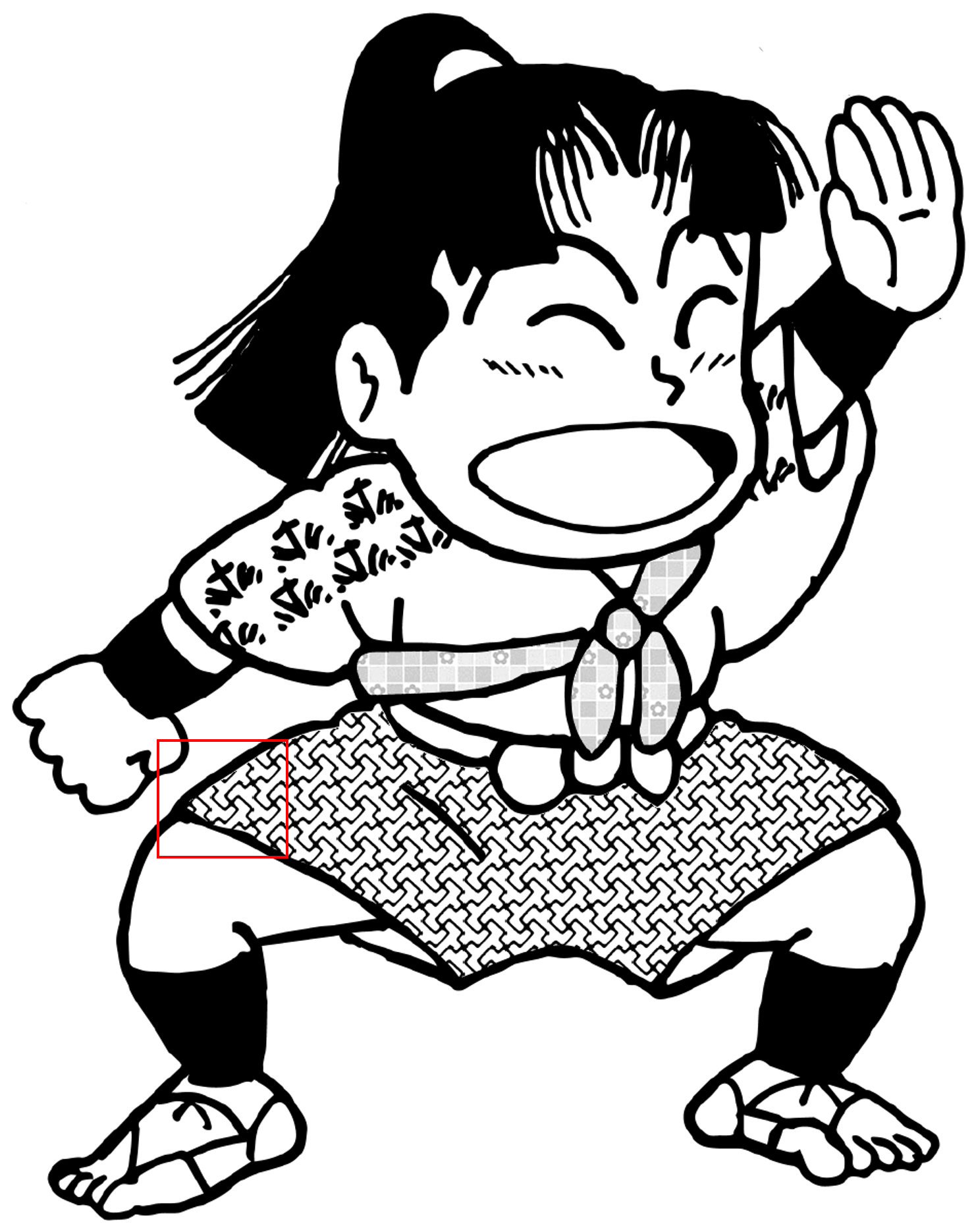}
        \includegraphics[scale=0.08]{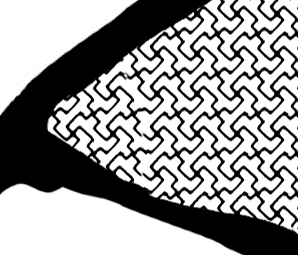}
    }
    \caption{Super resolution results about complex screentone. The last on the left is the size definition of different type of screentone. The second on the left is the area in the red box of LR image, and the last on the right is the area in the red box of SR image}
    \label{CH5_Sec1_fig2}
\end{figure}

To reach our goal, we trained the model with image filled with different density and direction of screentone in each pairs of training data. Through the deep neural network and different loss function, the model is capable to restored the LR image with the same density of screentone.
In this chapter, we will discuss our research results. We will illustrate all LR images as well as their magnified results restored by our system, compare the results and analyze screentone density. For dotted, linear, and gridded screentone, we adopt the methodology in Yao's work~\cite{7399427} for a comparative analysis of dot size, spacing, and angle as well as line width and spacing. Meanwhile, considering difficulties in mathematically defining complex screentone, our approach to comparison was to manually designate a screentone’s component region and then calculate the average size of screentone-composing unit areas to verify whether screentone density from before and after magnification remains the same. Figure~\ref{CH5_Sec1_fig1} shows results of simple screentone like dotted, linear, and gridded screentone. We can easily define the density and direction of these screentones by measure the size of dot, width of line or angle line. Figure~\ref{CH5_Sec1_fig2} shows results of more complex screentone. For these screentone, we define the density by manually measure the size of each repeat pattern. The definition of the density are shown in each pair of data on last on the left in Figure~\ref{CH5_Sec1_fig2}.

\subsection{Screentone Classification Retults}
We compare the results between our method and Yao's method~\cite{7399427}(Local Binary Pattern(LBP)). We use 23 types of screentone, including dotted, linear, gridded and other complex screentone for testing. The test results are shown in table~\ref{tabular1}. We use 100 images for testing in every row. The numbers in the table are the accuracy in percentage. The test data from the first row to the fourth row contains only a single type of screentone, and the fifth row contains multiple types of screestone in a single image. The left number in the parenthesis of forth row(75) represent the proportion of the the LBP method successfully classify the screentone as complex screentone; the right number(10) represent the proportion of the the LBP method successfully classify which class of screentone it is. LBP method cannot work on an image containing multiple type of screentone, so we only test these image using our method in the fifth row.

\begin{table}[htbp]
    \begin{center}
        \begin{tabular}{|c|c|c|} 
        \hline 
        &LBP&Ours\\
        \hline  
        Single-Dot&100&100\\
        \hline 
        Single-Stripe&92&100\\
        \hline 
        Single-Grid&93&99\\
        \hline 
        Single-Others&(75,10)&96\\
        \hline 
        Mix Screentones&-&97.4\\
        \hline 
        \end{tabular}
        \vspace{0.6mm}
    
    \caption{The results of using the Local Binary Pattern (LBP)) and using the semantic segmentation to classify types of screentone. }
    \label{tabular1}
    \end{center}
\end{table}

\subsection{Structure Network Retults}
We Fine-tuned ESRGAN and compare the results with our results after adding structure net. ESRGAN tends to generate artifacts when restoring some structural screentone, leading to poor results. We compare the peak signal-to-noise ratio (PSNR) and the structural similarity (SSIM) of such images with the results of our method. We take the average of 100 images, the results are shown in Table ~\ref{tabular2}. It can be seen that our method can achieve better results numerically.

\begin{table}[htbp]
    \begin{center}
        \begin{tabular}{|c|c|c|} 
        \hline 
        &ESRGAN&Ours\\
        \hline  
        PSNR&13.9216&14.9772\\
        \hline 
        SSIM&0.7053&0.8582\\
        \hline 
        \end{tabular}
        \vspace{0.6mm}
    
    \caption{COmparison between ESRGAN and our method in image with strictural screentone.}
    \label{tabular2}
    \end{center}
\end{table}

\section{Conclusion and Future works}
We improved ESRGAN \cite{ESRGAN} to maintain the style and density of screentones while enhance the resolution of manga images, thereby preserving the meaning of screentones in manga. To achieve this, we first extracted and classified screentones using segmentation\cite{DBLP:journals/corr/RonnebergerFB15}. Then, we applied different pre-trained super-resolution networks according to the classification to avoid interference between different types of screentones. We also added a loss function to maintain the structure of screentones and prevent artifacts from occurring. Finally, we directly applied super-resolution to the original image and pasted each screentone block that has been enhanced by super-resolution back into the image. Although different types of screentone SR networks need to be trained beforehand, our method reduces the need for manual parameter adjustment compared to vectorization methods and supports more complex styles.

\begin{figure}[htbp]
  \begin{center}
    \includegraphics[scale=0.25]{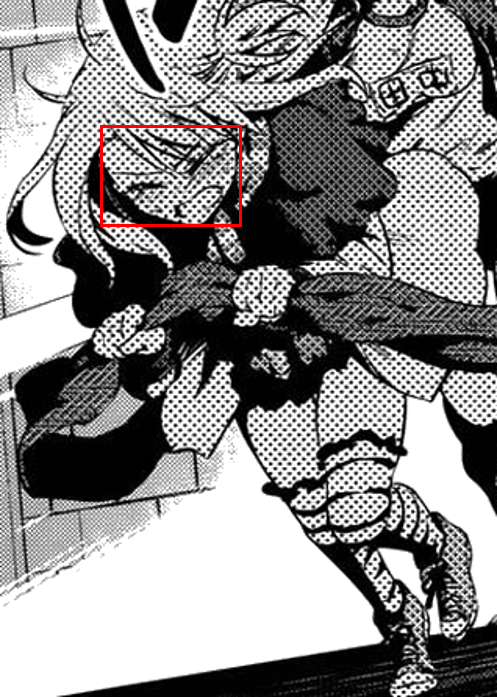}
    \includegraphics[scale=0.5]{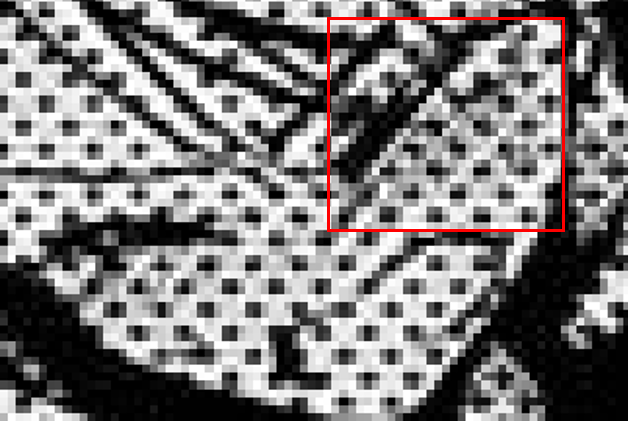}
    \caption{The blending of the sreentones is hard to classify, and super-resolution method can't preserve all the detail, the area finally distinguish as a dot screentone and the shading of the red cheek might disappear.}
    \label{CH6_fig1}
  \end{center}
\end{figure}

As the generation of upscale screentone requires training, styles that have not been trained before cannot be subjected to super-resolution. Additionally, in some cases, two different types of screentone may be blended to express different material effects or emotions, such as in the cheek area of the red box in Figure ~\ref{CH6_fig1}. Although the human can recognize the dot type as points, it will confuse the recognition model and also affect the results of super-resolution. In the future, we may contaminate the dot area in the training data to teach the model to have the ability to handle such situations.

\bibliographystyle{IEEEtran}
\bibliography{reference}

\newpage

\section{Biography Section}
 
\vspace{11pt}

\begin{IEEEbiography}[{\includegraphics[width=1in,height=1.25in,clip,keepaspectratio]{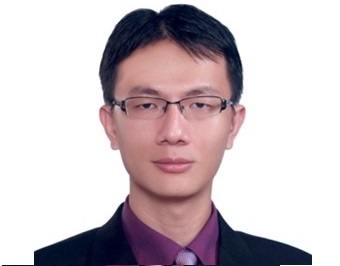}}]{Chih-Yuan Yao}
received the M.S. and Ph.D. degrees in computer science and information engineering from National Cheng-Kung University, Tainan, Taiwan, in 2003 and 2010, respectively. He is an Associate Professor with the Department of Computer
Science and Information Engineering, National Taiwan University of Science and Technology (NTUST),Taipei, Taiwan. His research interest include computer graphics, including mesh processing and modeling, and non-photorealistic rendering (NPR)
\end{IEEEbiography}

\begin{IEEEbiography}[{\includegraphics[width=1in,height=1.25in,clip,keepaspectratio]{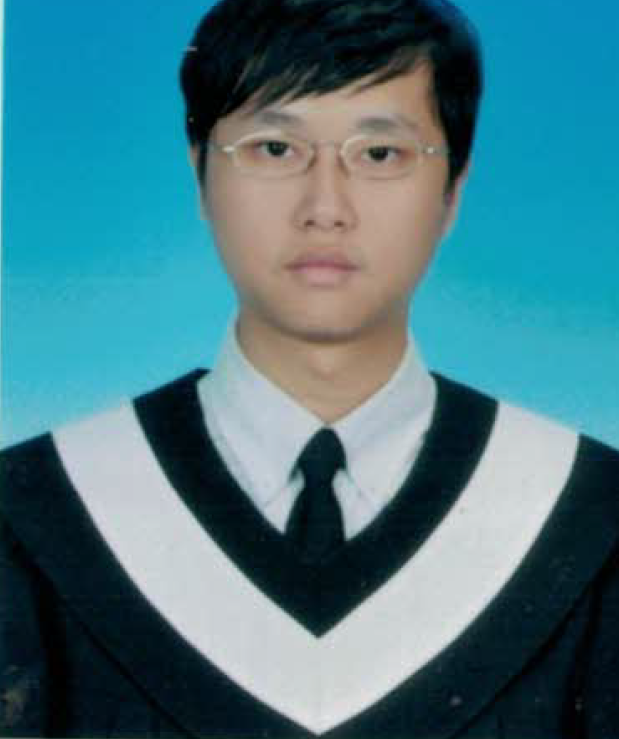}}]{Husan-Ting Chou}
received the B.S. degree from the Department of Electronic and Computer Engineering, National Taiwan University of Science and Technology (NTUST), Taipei, R.O.C., in 2013
and the M.S degree from the Department of Computer Science and Information Engineering, National Taiwan University of Science and Technology, Taipei, Taiwan, in 2015. 
He is currently pursuing the Ph.D degree with the Department of Computer Science and Information Engineering.
His research interests include graphics, vision, and multimedia.
\end{IEEEbiography}

\begin{IEEEbiography}[{\includegraphics[width=1in,height=1.25in,clip,keepaspectratio]{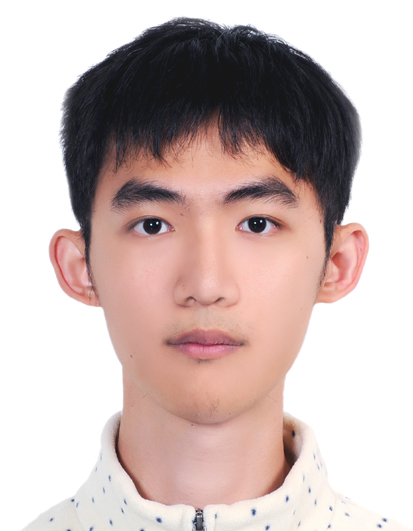}}]{Yu-Sheng Lin}
received the M.S degree from the Department of Computer Science and Information Engineering, National Taiwan University of Science and Technology, Taipei, Taiwan, in 	2021. His research interests include graphics, super resolution.
\end{IEEEbiography}

\begin{IEEEbiography}[{\includegraphics[width=1in,height=1.25in,clip,keepaspectratio]{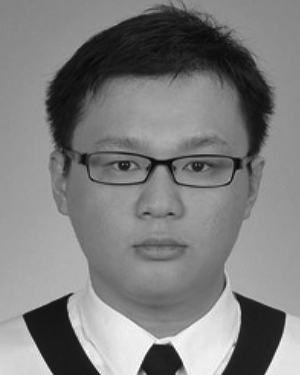}}]{Kuo-wei Chen}
received the B.S. degree from the Department of Electronic and Computer Engineering, National Taiwan University of Science and Technology (NTUST), Taipei, Taiwan, R.O.C. in 2013, and the M.S. degree from the Department of Computer Science and Information Engineering, NTUST, Taipei, Taiwan, R.O.C., in 2015. He is currently working toward the Ph.D. degree at the Department of Computer Science and Information Engineering, NTUST.
His research interests include the area of graphics, vision, and multimedia.
\end{IEEEbiography}

\vspace{11pt}


\vfill

\end{document}